\documentclass[10pt,twocolumn,letterpaper]{article}

\usepackage{iccv}
\usepackage{times}
\usepackage{epsfig}
\usepackage{graphicx}
\usepackage{amsmath,amssymb}
\usepackage{subfigure}
\usepackage{epstopdf}
\usepackage{makecell}
\usepackage{arydshln}
\usepackage{cite}
\usepackage{booktabs}
\usepackage{hhline}
\usepackage{textcomp,booktabs}
\usepackage[usenames,dvipsnames]{color}
\usepackage{colortbl}
\usepackage{color}
\usepackage[ruled]{algorithm2e}

\usepackage[pagebackref=true,breaklinks=true,letterpaper=true,colorlinks,bookmarks=false]{hyperref}

 \iccvfinalcopy 

\def\iccvPaperID{9551} 

\begin{document}

\title{Unsupervised Outlier Detection using Memory and Contrastive Learning }
\author{Ning Huyan{\small\(^{1}\)}
	\and
	Dou Quan{\small\(^{1}\)}
	\and
	Xiangrong Zhang\small{\(^{*1}\)}
	\and
	Xuefeng Liang\small{\(^{1}\)}
	\and
	Jocelyn Chanussot\small{\(^{2}\)}
	\and
	Licheng Jiao{\small\(^{1}\)}\\
	{\small\(^{1}\)}School of Artificial Intelligence, Xidian University, Shaanxi, China \\ 
	
	{\small\(^{2}\)} Research center of Inria Grenoble-Rhone-Alpes, France\\
	{\tt\small xrzhang@mail.xidian.edu.cn, xliang@xidian.edu.cn, dquan@stu.xidian.edu.cn}
}
\maketitle


\ificcvfinal\thispagestyle{empty}\fi

\begin{abstract}
Outlier detection is one of the most important processes taken to create good, reliable data in machine learning. The most methods of outlier detection leverage an auxiliary reconstruction task by assuming that outliers are more difficult to be recovered than normal samples (inliers). However, it is not always true, especially for auto-encoder (AE) based models. They may recover certain outliers even outliers are not in the training data, because they do not constrain the feature learning. Instead, we think outlier detection can be done in the feature space by measuring the feature distance between outliers and inliers. We then propose a framework, MCOD, using a memory module and a contrastive learning module. The memory module constrains the consistency of features, which represent the normal data. The contrastive learning module learns more discriminating features, which boost the distinction between outliers and inliers. Extensive experiments on four benchmark datasets show that our proposed MCOD achieves a considerable performance and outperforms nine state-of-the-art methods.
\end{abstract}

\section{Introduction}
Outliers, also named anomalies, are unknown samples that deviate significantly from the distribution of normal samples (inliers). Outliers could be the lesions in medical images~\cite{schlegl2017unsupervised}, frauds of banking transactions~\cite{2018A}, and abnormal behavior in video surveillance~\cite{wang2020cluster, sabokrou2018adversarially}. Detecting outliers is a crucial issue and important in everyday life. Since outliers rarely appear and are unpredictable, outliers and inliers are commonly unlabeled. The tasks of outlier detection (OD) in practice are unsupervised, which aims at identifying the commonalities within the unlabeled data to facilitate outlier detection~\cite{Chalapathy2019Deep}. The challenge of OD is that the data distribution of inliers and outliers can be extremely complicated, especially, they are unlabeled~\cite{liang2017enhancing}.

\begin{figure}[!tp]
	\centering {\includegraphics[width=1\linewidth]{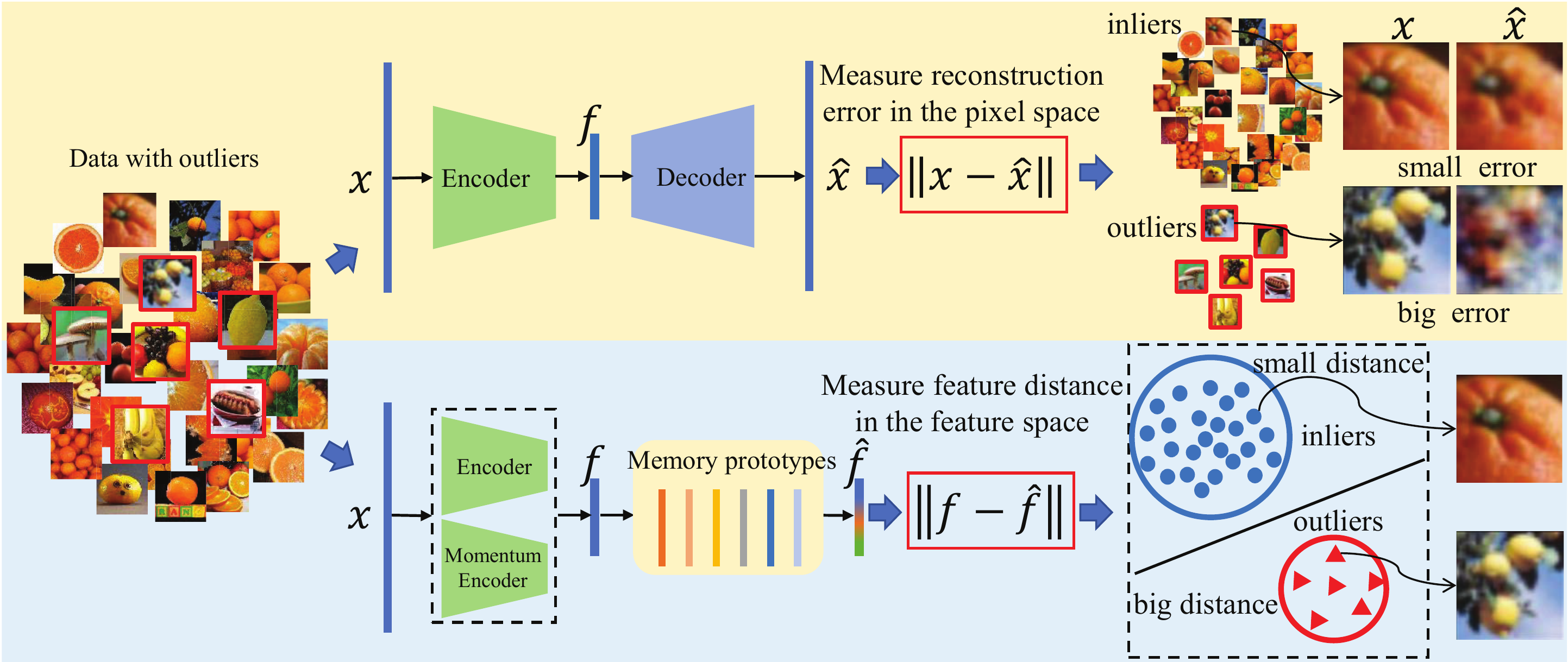}}
	\caption{\footnotesize{The upper one is an AE based module, which detects outliers according to the reconstruction error in the pixel space. The lower one is our MCOD, which detects outliers in the feature space using self-supervised contrastive learning and a memory model.}}
	\label{fig1}
\end{figure}

In the era of deep learning, many studies tried to leverage an auxiliary deep reconstruction model for OD. In particular, auto-encoder (AE)~\cite{bourlard1988auto} is a widely used model~\cite{xia2015learning, perera2019ocgan}, as shown in Fig. \ref{fig1}, AE based methods determine if a sample is outlier or not according to the reconstruction error in the pixel-space, because they assume that inliers are easier to be reconstructed than outliers. However, it is not always true. The reason behind this fact is that AE models mainly focus on the quality of sample reconstruction, do not care about if the encoded features solely represent the inlier rather than outliers. Specifically, most AE based methods do not constrain the consistency of encoded feature for intra-class samples and the discrimination for inter-class samples during training process. Consistency denotes the features of samples in the same class are close, and discrimination means the features in different classes are distinct. Due to the lack of these two constraints, the encoded feature in AE may include the shared features between inliers and outliers that induce a small reconstruction error for outliers. The research~\cite{perera2019ocgan} that also reported the outliers could be well reconstructed, even if the model was only trained on inliers.

\begin{figure*}[!tp]
	\centering
	{\includegraphics[width=0.9\linewidth]{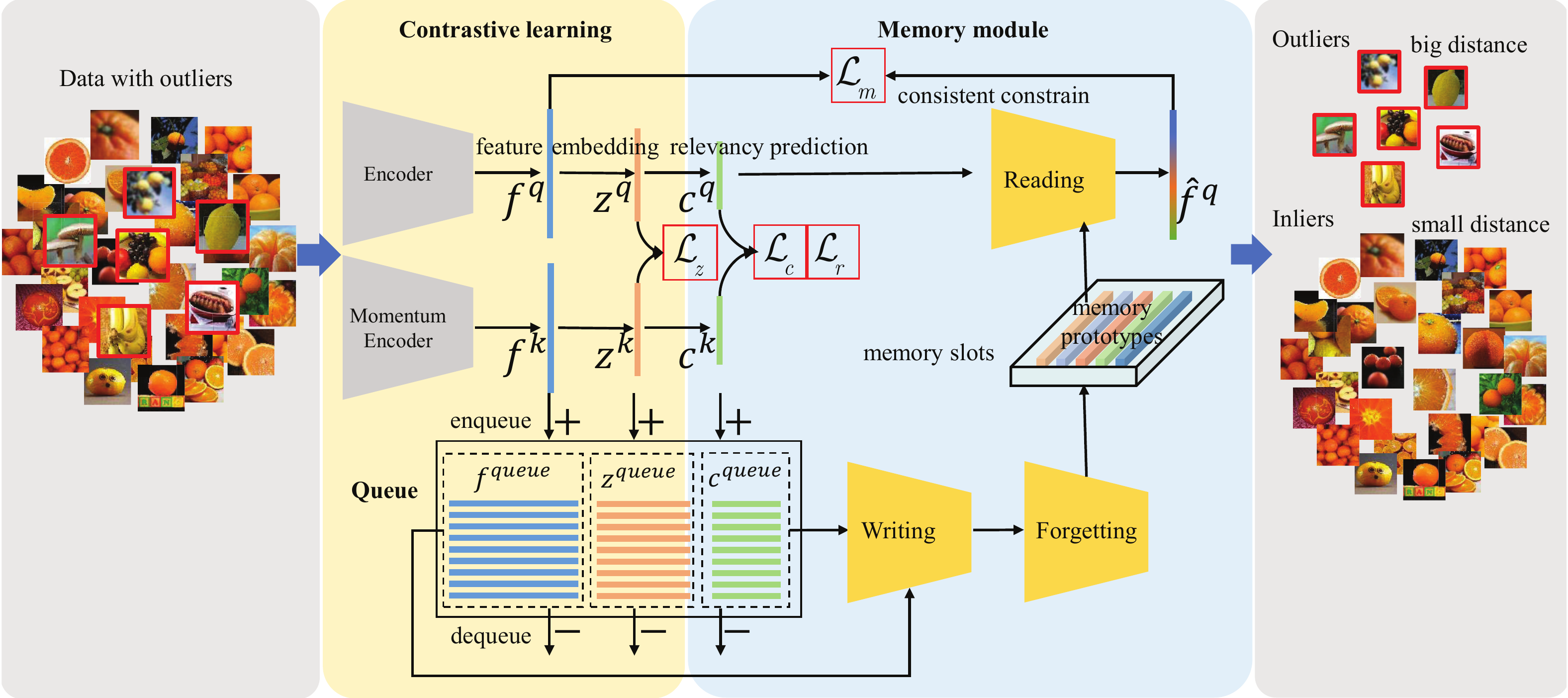}}
	\caption{\footnotesize{Schematic illustration of MCOD. It consists of two symmetrical encoders, one queue, memory slots, and three operations. $f$, $z$, $c$ are outputs of feature layer, embedding layer and relevancy prediction layer, respectively. In the training process, two encoders extract features from the input sample and its augmentation and perform contrastive learning. The reading operation selects the corresponding memory prototypes for input samples, and the writing operation updates memory prototypes. The forgetting operation adds perturbations to prototypes to suppress the outlier prototypes in memory slots.}}
	\label{framework}
\end{figure*}

In our study, we find that the outlier can be detected in the feature space, not necessary to leverage the reconstruction model. To this end, the extracted features must be consistent for inliers, meanwhile, discriminating to separate outliers from inliers. Following this idea, we propose an unsupervised outlier detection method using memory and contrastive learning (MCOD) shown in Fig. \ref{framework}. Firstly, we propose a memory module~\cite{graves2014neural,gong2019memorizing, park2020learning} to learn the consistent features of inliers. It has three operations (writing, forgetting, and reading) and a memory slot. The memory slots store the memory prototype of each class, which is the predicted cluster center of the class. The writing computes memory prototypes and updates them into memory slots. More importantly, we propose a forgetting operation that adds perturbations to prototypes to suppress the outlier prototypes in memory slots. The reading constrains the feature of sample to be close to its corresponding prototype. Therefore, MCOD is able to learn the consistent features. Secondly, we design a self-supervised contrastive learning module to learn the discriminating features from unlabeled inliers and outliers. It has two symmetrical encoders in which the sample label is obtained by self-augmentation.

The main contributions of this paper are concluded as follows:
\begin{itemize}
	\item We propose an outlier detector framework (MCOD), which learns the discriminating feature by means of self-supervised contrastive learning and the consistent feature using memory mechanism.
	\vspace{-1mm}
	\item We conduct the outlier detection in the feature space without reconstructing the original samples.
	\vspace{-1mm}
	\item MCOD is trained by an end-to-end mode. Extensive experiments on four benchmarks demonstrate that MCOD outperforms nine SOTA methods.
\end{itemize}

\section{Related work}

\textbf{Outlier detection.} The mainstream approaches of outlier detection (OD) are unsupervised frameworks, which assume that outliers are minority and significantly deviated from inliers~\cite{Chalapathy2019Deep}. Generally, they can be grouped into two categories: \textit{discrimination-based} and \textit{reconstruction-based} methods. The discrimination-based OD methods detect outliers according to the hyperplane or hypersphere in the feature space, in which inliers are distributed in the hypersphere and outliers are outside. The representative methods are one-class SVM (OC-SVM)~\cite{scholkopf1999support}, support vector data descriptor (SVDD)~\cite{TaxSVDD} and the deep version of SVDD (deep SVDD)~\cite{pmlr-v80-ruff18a}, which detect outliers by minimizing the volume of the hypersphere of the inliers. However, they have less ability of learning the distribution of complex high-dimensional data.

The reconstruction-based OD methods often employ auto-encoders to detect outliers by measuring the reconstruction errors ~\cite{xia2015learning, zhou2017anomaly, zong2018deep, lai2019robust, eduardo2020robust}. They think that outliers are more difficult to be reconstructed than inliers because outliers are the minority. Zong et al.~\cite{zong2018deep} integrated AE and a density estimation model based on Gaussian Mixture Model (GMM) for a better inlier reconstruction. Zhou et al. ~\cite{zhou2017anomaly} proposed a deep auto-encoder (RDAE) associated with the robust principal component analysis (RPCA) to discard the outlier in the feature space. Lai et al.~\cite{lai2019robust} followed this idea and designed a robust subspace recovery layer (RSR) for AE. Unlike these methods, the work~\cite{xia2015learning} gradually separated inliers and outliers by alternatively minimizing reconstruction errors according to labeled inliers. All above methods made great efforts to achieve a better reconstruction, even relied on inliers only for training. Nevertheless, they neglected to constrain the features to solely represent inliers. Therefore, such AE based methods may reconstruct outliers to some extent.

\textbf{Self-supervised contrastive learning.} Self-supervised method is able to learn feature representation of unlabeled data by means of supervision. The method~\cite{wangself} applied the self-supervised learning to build pseudo labels using different pretext tasks and utilized the uncertainty of discriminating model to detect outliers. Instead of matching predictions to pseudo labels, contrastive learning~\cite{chopra2005learning, he2020momentum, jing2020self} both maximizes the agreement between a sample and its augmentations, and the differences between different samples. To effectively enlarge the number of negative samples and maintain consistency of samples across preceding several preceding mini-batches, MoCo~\cite{he2020momentum} proposed a dynamic dictionary as a queue that enqueue current mini-batch samples coming from an updated encoder and dequeue the earliest sample. In this work, we apply contrastive leaning and queue to learn the discriminating features of unlabeled data, which facilitates cluster predictions for subsequently memory module learning~\cite{zhong2020deep}.

\textbf{Memory mechanism.} Recently, memory mechanism has demonstrated a remarkable effectiveness on many tasks~\cite{santoro2016one, zhu2019dm, lee2018memory}. It has also been applied for anomaly detection problems~\cite{park2020learning, gong2019memorizing}. However, their feature learning is based on auto-encoder and semi-supervised architecture, whose training data are inliers only. With appropriate reading and writing operations, the memory module is able to learn the prototypes that represent inliers and ignore outliers. Due to this property, we leverage memory mechanism to achieve the consistent features from data mixed with unlabeled inliers and outliers.

\section{The proposed framework}

The framework of our proposed unsupervised outlier detector MCOD is shown in Fig. \ref{framework}. The contrastive learning takes charge of learning the discriminating features. The memory module is responsible for learning the consistent prototypes of inliers. More details are given in the following sections.

\subsection{Problem formulation}
Let us consider a dataset $X \subset \mathcal{X}$, where $\mathcal{X}$ is the data space spanned by all samples in the space. The dataset is mixed up with both inliers $X_{in}$ and outliers $X_{out}$, and satisfy $X = {X_{in}} \cup {X_{out}}$ and ${X_{in}} \cap {X_{out}} = \emptyset $. The number of outliers $X_{out}$ is much smaller than the number of $X_{in}$. The goal of outlier detection is to separate $X_{out}$ from data $X$. Generally, the feature of outliers $X_{out}$ should be out of the feature distribution of inliers $X_{in}$. Thus, it is reasonable to detect outliers in the feature space. Our goal is to design a framework to learn the consistent feature of the entire data distribution to represent inliers, and the discriminating features to separate outliers from inliers.

\subsection{Contrastive learning for discriminating feature}

Since the data are unlabeled, we leverage a self-supervised contrastive learning to learn the discriminating features. Usually, contrastive learning framework has two encoders, which are contrastively optimized by sample pairs. The framework tries to maximize the feature similarity of positives (a sample and its self-argumentations) and the feature difference of negatives (two different samples).

In MCOD, a sample is augmented by various transformations, including random flipping, random rotating, random crop, random contrast changes, and Gaussian blur. Then, we have a self-labeled pair $(x_i^q, x_i^k)$. We design two symmetrical encoders as shown in the contrastive learning block of Fig. \ref{framework}. The input of the first encoder ${E_q}$ is $x_i^q$, and the input of the second encoder ${E_k}$ is $x_i^k$. The features $f_i^q$ and $f_i^k$ of the sample-pair are formulated as
\begin{equation}\label{}
\begin{aligned}
f_i^q=E_q(x_i^q), \quad f_i^k=E_k(x_i^k),
\end{aligned}
\end{equation}

To keep more information in features, we build a non-liner embedding layer $g_q$ after encoders $E_q$, and then conduct contrastive learning using these embedding representations,
\begin{equation}\label{}
\begin{aligned}
z_i^q=g_q(f_i^q),\quad
z_i^k=g_k(f_i^k),
\end{aligned}
\end{equation}
where $z_i^q$ and $z_i^q$ are the embedding representations of image $x_i^q$ and $x_i^k$ respectively.

To better optimize the embedding representations from contrastive learning, we follow the idea of~\cite{he2020momentum} that expands the set of negative samples from preceding mini-batches by create a queue. The queue has a capacity of 4096, and will be filled up after first few iterations. Afterwards, the new embeddings in the later mini-batches are added to the queue (enqueue), meanwhile, the earliest embeddings are removed (dequeue). To optimize the embedding representations from contrastive learning, we adopt the InfoNCE loss~\cite{chen2020improved, he2020momentum, oord2018representation}

\begin{equation}
{\mathcal{L}_z} =-\frac{1}{N}\sum\limits_{i = 1}^N {\log \frac{{\exp ({{z_i^q \cdot z_i^k} \mathord{\left/
 {\vphantom {{z_i^q \cdot z_i^k} {{\tau _z}}}} \right.
 \kern-\nulldelimiterspace} {{\tau _z}}})}}{{\sum\limits_{z_j^k \in \mathcal{Z}} {\exp ({{z_i^q \cdot z_j^k} \mathord{\left/
 {\vphantom {{z_i^q \cdot z_j^k} {{\tau _z}}}} \right.
 \kern-\nulldelimiterspace} {{\tau _z}}})} }}},
\label{feature_infoNCE}
\end{equation}
where $N$ indicates the size of mini-batch, $\mathcal{Z} = \{{z_i}^k\} \cup \{z\}$, ${z}$ represents embedding stored in the queue, $\tau_z$ is the temperature parameter.
\subsection{Memory module for consistent feature}
As outliers are significantly deviated from inliers, they should be separated according to the distance in the feature space. We need a consistent feature to represent inliers, which can be considered as the cluster center of sample features in one class, called prototypes. Therefore, we design a memory module which maintain explicit memory slots to store the renewable prototypes as shown in the memory model block of Fig. \ref{framework}. The features of inputs are constrained to be similar with their corresponding prototypes.

\textbf{Memory relevancy prediction.} To compute a memory prototype, we need to predict the relationship between a sample and each cluster center. Then, we add a clustering layer, $h(\cdot)$, after the embedding layer, $g(\cdot)$, which is a fully connected layer for relation estimation. Its output is a probability,

\begin{equation}\label{}
\begin{aligned}
c_i^q=h_q(z_i^q),\quad
c_i^k=h_k(z_i^k),
\end{aligned}
\end{equation}
where $c^q_i =\{c_{ij}^q\}$ and $c^k_i =\{c_{ij}^k\}$, $j = 1 \ldots K$ represent the probabilities of the samples $x_i^q$ and $x_i^k$ belonging to the $j_{th}$ memory prototype, respectively. We call them as the memory relevancy prediction. $K$ is the number of memory prototypes. A sample is expected to be close to its corresponding prototype, but distant from other prototypes. In other words, the sample should locate in one cluster, but not other clusters. This can be achieved by InfoNCE loss using $K$ predictions as samples,
\begin{equation}
{\mathcal{L}_c} = -\frac{1}{K}\sum\limits_{i = 1}^K {{ \log  \frac{{\exp ({{v_i^q \cdot v_i^k} \mathord{\left/
 {\vphantom {{v_i^q \cdot v_i^k} {{\tau _c}}}} \right.
 \kern-\nulldelimiterspace} {{\tau _c}}})}}{{\sum\limits_{j = 1}^K {\exp ({{v_i^q \cdot v_j^k} \mathord{\left/
 {\vphantom {{v_i^q \cdot v_j^k} {{\tau _c}}}} \right.
 \kern-\nulldelimiterspace} {{\tau _c}}})} }}}},
\label{cluster_infoNCE}
\end{equation}
where $v_i^q \in \{{c_{\cdot j} ^q}^{\top}\}$, $v_i^k \in \{{c_{\cdot j} ^k}^{\top}\}$, $j = 1 \ldots K$ . $\tau _c$ is a temperature parameter. $v_i^q$ and $v_i^k$ are normalized before the optimization.

Moreover, we add a regularization to prevent the model falling into a local minimum where the majority gathers to the minority clusters~\cite{zhong2020deep}. It is formulated as
\begin{equation}
{\mathcal{L}_r} = {1 \over N}\sum\limits_{j = 1}^K {{{\left( {\sum\limits_{i = 1}^N {{c_{ij}^q}} } \right)}^2}},
\label{regularization}
\end{equation}

\textbf{Memory reading.} To ensure the consistence of features, we need to minimize the distance between the sample feature and its corresponding memory prototype. To this end, we apply a reading operation with assistance of the memory relevancy prediction.

Given memory prototypes  $M=\{m_1, \dots, m_j\}\in \mathcal{R}^{K \times d}$, finding the most similar prototype can be achieved by a hard assignment memory reading $r(\cdot)$,

\begin{equation}\label{}
\hat f_i^q = r({c^q_{ij}, m_j}) = \mathop {\arg \max }\limits_{{m_j} \in M} ({c^q_{ij}}),
\end{equation}
where ${\hat f}_i^q$ is the read memory prototype of sample $x_i^q$ from memory slots, and $c_{ij}^q$ is the memory relevancy prediction between $f_i^q$ and $m_j$. Considering that samples usually share certain features across a few relevant clusters, we adopt a soft assignment memory reading,
\begin{equation}
{\hat f}_i^q  = r({c^q_{ij}, m_j}) = \sum\limits_{j = 1}^K {\varphi ({c^q_{ij}}){m_j}},
\label{read}
\end{equation}
where $\varphi ( \cdot )$ is the softmax function.

In training process, our goal is to minimize the distance between $f_i^q$ and ${\hat f}_i^q$ under current memory prototypes and cluster prediction. The feature consistency loss based on the Euclidean distance is
\begin{equation}\label{}
{\mathcal{L}_m} = {1 \over N}\sum\limits_{i = 1}^N || {\hat f}_i^q - f_i^q||^2,
\label{reconstruction_loss}
\end{equation}
By this way, $\mathcal{L}_m$ can push $f_i^q$ closer to its corresponding memory prototype.

\textbf{Memory writing.} To find better consistent features, memory prototypes should be updated during the training. Thus, we employ memory writing operation to generate new memory prototypes according to the features and their cluster predictions stored in the queue. The writing operation, $w(\cdot)$, is
\begin{equation}
{m_j} = w(f_i^{queue}, c_{ij}^{queue}) = \sum\limits_{i = 1}^{N_Q} {\varphi (c_{ij}^{queue})f_i^{queue}},
\label{write}
\end{equation}
where $N_Q=4096$ is the number of features stored in the queue. We update the memory in the queue at each training iteration, which can keep memory prototypes more stable.

\textbf{Memory forgetting.} Due to the training data are mixed with outliers, the learned memory prototypes may also include outlier features and result in a worse performance. Then, we propose a memory forgetting operator to add perturbations to the memory prototypes using the attention mechanism. The attention weight of each memory prototype is determined by the number of related support samples.
\begin{equation}
{n_j} = \sum\limits_{i = 1}^{{N_Q}} {{\mathbb{I}_{\max }}({c_{ij}})},
\end{equation}
where $\mathbb{I}_{\max}(\cdot)$ is $1$ when $c_{ij}$ is the maximal output value for sample $x_i$ , otherwise is $0$. $n_j$ is the number of the support samples of the $j_{th}$ memory prototype, and $n_1+n_2+\dots+n_K=N_Q$. The smaller $n_j$ is, the more likely the prototype belongs to outliers. Thus, we add the Gaussian noise to suppress the these prototypes in the memory slots. The forgetting operation is formulated as
\begin{equation}\label{}
\begin{aligned}
m_j = m_j + {\eta}_j,
\label{forgetting_eq}
\end{aligned}
\end{equation}
where ${\eta}_j \sim \mathcal{N}(0, \sigma_j)$ and $\sigma_j = 1- n_j /N_Q$.

The memory forgetting operation can force the outliers to keep distant from memory prototypes.

\textbf{Learning and inference.} The overall objective function of MCOD can be formulated as
\begin{equation}
\mathcal{L} = \mathcal{L}_z + \mathcal{L}_c + \mathcal{L}_m + \lambda \mathcal{L}_r,
\end{equation}
where $\lambda$ is a weight parameter of $\mathcal{L}_r$.

In the training process, $E_q$ plays as a general network, and $E_k$ plays as a a momentum network. Their parameters are gradually updated as

\begin{equation}
{\theta ^k}: = \alpha {\theta ^k} + (1 - \alpha ){\theta ^q}.
\end{equation}
where $\theta^q$ and $\theta^k$ are the parameters set of ($E_q$, $g_q$, $h_q$) and ($E_k$, $g_k$, $h_k$) respectively. $\alpha \in [0,1)$ is momentum coefficient which is empirically set to be $\alpha = 0.999$ in all our experiment. The training procedure of MCOD is summarized in Algorithm \ref{algorithm}.

In the inference process, the reading operation selects the corresponding memory prototypes of samples. The outliers can be detected according to the distance between their feature and the read memory prototype.

\begin{algorithm}[!htb]
	\caption{Training algorithm of MCOD}
	\label{algorithm}
	\KwIn{Training samples $X \in \mathcal{X}$, the number of memory prototypes $K$, temperature parameters $\tau_z$ and $\tau_c$}
	\KwOut{A trained MCOD with parameters sets $\Theta $}
	\For{each iteration}{
		\textbf{Step1:} Choose a random mini-batch of samples and generate augmentations for the samples;
		
		\textbf{Step2:} Calcuate $\mathcal{L}_z$, $\mathcal{L}_c$ and $\mathcal{L}_r$ by the outputs $z$ and $c$ using Eq. \ref{feature_infoNCE}, Eq. \ref{cluster_infoNCE} and Eq. \ref{regularization};
		
		\textbf{Step3:} Add current $f_q$, $z_q$ and $c_q$ into the queue and remove the earliest ones;
		
		\textbf{Step4:} Conduct reading operation $r({\cdot})$ based on current memory prototypes $M$ using Eq. \ref{read} and calculate $\mathcal{L}_m$ in Eq. \ref{reconstruction_loss};
		
		\textbf{Step5:} Conduct writting operation $w({\cdot})$ using Eq. \ref{write} to update memory prototyes $M$;
		
		\textbf{Step6:} Conduct forgetting operation defined in Eq.\ref{forgetting_eq}.
	}
	
\end{algorithm}

\section{Experiment}

\begin{table*}[!htb]
	\footnotesize
	\centering
	\renewcommand\tabcolsep{3.0pt}
	\caption {\footnotesize{Comparison with nine SOTA methods on four datasets with two different proportions of outliers ($p=0.1$ and $p=0.2$). The best performance is shown in bold font.}}
	\begin{tabular}{lccccccccccl}
		\Xhline{0.5pt}\Xhline{0.5pt}
		Dataset&$p$&CAE\cite{masci2011stacked}& DRAE\cite{xia2015learning}& DSEBM\cite{zhai2016deep}& RDAE\cite{zhou2017anomaly}&MOGAL\cite{liu2019generative}& RSRAE\cite{lai2019robust}& RSRAE+\cite{lai2019robust}&SSD-IF\cite{wangself}& E3Out\cite{wangself}&MCOD \\
		\Xhline{0.5pt}
		\multicolumn{12}{c}{AUROC\%}\\
		\Xhline{0.5pt}
		MNIST&0.1&68.0$\pm$0.7&  66.9$\pm$2.6& 60.5$\pm$8.5 &71.8$\pm$1.0& 30.9$\pm$1.4 &84.8$\pm$0.9& 73.9$\pm$ 1.4&93.8$\pm 0.4$& 94.9$\pm$ 0.1&\textbf{96.2$\pm$2.1}\\
		&0.2&64.0$\pm$ 1.0& 67.3$\pm$2.1 &56.3$\pm$6.5 &67.0$\pm$0.7& 37.8$\pm$2.4 &78.9$\pm$1.7 &74.6$\pm$0.8&90.5$\pm$0.0& 92.9$\pm$0.1&\textbf{94.1$\pm$3.3}\\
		F-MNIST&0.1 & 70.3 $\pm$0.9 & 67.1$\pm$ 1.8 &53.2$\pm$ 3.0& 75.3$\pm$ 1.2 & 22.8$\pm$ 1.3 &78.3$\pm$2.0& 84.6$\pm$ 1.4 &91.0$\pm$ 0.0&93.5$\pm$ 0.1&\textbf{98.7$\pm$0.6}\\
		
		&0.2&64.4$\pm$ 1.8 & 65.7$\pm$ 1.9& 53.1$\pm$ 5.6& 70.9$\pm$ 1.1&34.0$\pm$ 2.9& 74.5$\pm$3.1 &82.1$\pm$ 1.3 &87.7$\pm$ 0.0&92.1$\pm$ 0.1&\textbf{94.8$\pm$0.9}\\
		CIFAR10&0.1&55.8$\pm$ 1.4 &56.0$\pm$ 0.1& 60.2$\pm$ 1.2& 55.4$\pm$ 0.6 &56.2$\pm$ 0.7& 56.6$\pm$1.1& 55.5$\pm$ 2.6&64.3$\pm$ 0.0& 84.7$\pm$ 0.2&\textbf{87.6$\pm$ 4.1}\\
		
		&0.2& 54.7$\pm$ 1.2&  55.6 $\pm$ 0.3 &61.4$\pm$ 0.4& 54.2$\pm$ 0.9& 55.7$\pm$ 1.3& 55.6 $\pm$ 1.7& 57.4$\pm$ 1.2&58.9$\pm$ 0.1& 80.3$\pm$ 0.2&\textbf{85.3$\pm$1.7}\\
		CIFAR100&0.1& 51.2$\pm$ 0.4& 51.0$\pm$ 0.6 &59.2$\pm$ 1.0 &55.6$\pm$ 0.8& 53.2$\pm$ 1.0& 57.1$\pm$0.8& 57.7$\pm$ 0.5&56.4$\pm$ 0.0& 81.3$\pm$ 0.2&\textbf{87.7$\pm$ 4.2}\\
		
		&0.2&54.4$\pm$ 0.9& 55.5$\pm$ 0.3& 57.9$\pm$ 1.0& 54.9$\pm$ 0.6 &53.1$\pm$ 1.5 &56.3$\pm$0.7& 56.2$\pm$ 0.8 &53.2$\pm$ 0.1&79.1$\pm$ 0.2&\textbf{84.7$\pm$ 1.2}\\
		\Xhline{0.5pt}
		
		\multicolumn{12}{c}{AUPR-IN \%}\\
	
		\Xhline{0.5pt}
		MNIST&0.1& 92.0$\pm$ 0.3&  93.0$\pm$ 0.9 &91.6$\pm$ 2.3& 93.1$\pm$ 0.3 &78.8$\pm$ 3.4& 97.4$\pm$0.2& 95.7$\pm$ 0.4 &99.2$\pm$ 0.2&\textbf{99.4$\pm$ 0.0}&99.4$\pm$0.5\\
		&0.2& 82.7$\pm$ 0.5 & 86.6$\pm$ 1.2& 81.2$\pm$ 2.8& 89.2$\pm$ 0.7  & 70.6$\pm$ 1.2& 91.3$\pm$0.9& 91.1$\pm$ 0.3&97.3$\pm$ 0.0& \textbf{98.1$\pm$ 0.0}&97.9$\pm$1.4\\
		F-MNST&0.1 &94.3$\pm$ 0.3& 93.9$\pm$ 0.4& 88.9$\pm$ 0.8& 95.8$\pm$ 0.3 & 74.8$\pm$ 1.5 &96.2$\pm$0.4 &97.5$\pm$ 0.3&98.5$\pm$ 0.0& 99.0$\pm$ 0.0&\textbf{99.8$\pm$0.1}\\
		
		&0.2& 85.3$\pm$ 0.8 &86.9$\pm$ 0.9& 79.6$\pm$ 2.4 &89.2$\pm$ 0.7& 66.6$\pm$ 1.4& 90.4$\pm$1.2& 93.4$\pm$ 0.7 &95.5$\pm$ 0.0&97.4$\pm$ 0.1&\textbf{98.4$\pm$0.5}\\
		CIFAR10&0.1& 91.0$\pm$ 0.2& 90.7$\pm$ 0.1 &92.3$\pm$ 0.3& 90.7$\pm$ 0.2& 91.1$\pm$ 0.2& 91.4$\pm$0.2& 91.2$\pm$ 0.6&93.6$\pm$ 0.1& 97.7$\pm$ 0.0&\textbf{98.0$\pm$ 0.9}\\
		
		&0.2& 81.6$\pm$ 0.7& 81.7$\pm$ 0.2& 85.2$\pm$ 0.1& 81.0$\pm$ 0.4& 82.0$\pm$ 0.6 &82.1$\pm$0.8 &83.4$\pm$ 0.6 &84.1$\pm$ 0.0&93.5$\pm$ 0.1&\textbf{94.2$\pm$1.3}\\
		CIFAR100 &0.1& 90.3$\pm$ 0.1& 90.9$\pm$ 0.3 &92.2$\pm$ 0.3 &90.9$\pm$ 0.3& 90.4$\pm$ 0.4 &91.6$\pm$0.2 &91.9$\pm$ 0.1&91.6$\pm$ 0.0& 97.2$\pm$ 0.0&\textbf{98.1$\pm$ 0.8}\\
		&0.2&81.7$\pm$ 0.5 &81.8$\pm$ 0.2& 83.7$\pm$ 0.6& 81.5$\pm$ 0.3& 80.9$\pm$ 0.6& 82.7$\pm$0.3& 83.0$\pm$ 0.5&81.8$\pm$ 0.0& 93.1$\pm$ 0.1&\textbf{95.0$\pm$ 0.8}\\
		\Xhline{0.5pt}
		\multicolumn{12}{c}{AUPR-OUT \%}\\
		\Xhline{0.5pt}
		MNIST &0.1&32.9$\pm$ 0.7 &30.5$\pm$ 2.0 &23.0$\pm$ 5.3 &35.8$\pm$ 0.8 &15.2$\pm$ 3.8& 45.4$\pm$2.2& 30.6$\pm$ 1.5 &68.4$\pm$ 0.5&71.0$\pm$ 0.2&\textbf{85.5$\pm$5.1}\\
		
		&0.2& 40.7$\pm$ 1.0& 42.5$\pm$ 2.3& 32.3$\pm$ 6.0 &43.2$\pm$ 0.8 & 28.0$\pm$ 2.7& 53.0$\pm$1.6 &47.8$\pm$ 1.4 &70.1$\pm$ 0.0&76.3$\pm$ 0.1&\textbf{88.5$\pm$4.9}\\
		
		F-MNIST&0.1& 29.3$\pm$ 0.1&25.5$\pm$ 1.1& 19.7$\pm$ 2.3 &31.7$\pm$ 1.4  & 14.8$\pm$ 2.1& 37.0$\pm$ 4.5& 55.0$\pm$2.4 &70.1$\pm$ 0.0&77.2$\pm$ 0.1&\textbf{90.4$\pm$3.7}\\
		
		&0.2&36.8$\pm$ 1.3 & 36.6$\pm$ 2.5 &31.7$\pm$ 4.8& 41.4$\pm$ 0.9& 28.3$\pm$ 2.9 &46.3$\pm$5.1 &63.5$\pm$ 1.3 &71.5$\pm$ 0.0&82.0$\pm$ 0.2&\textbf{82.6$\pm$2.0}\\
		CIFAR10 &0.1&14.4$\pm$ 0.7 &14.7$\pm$ 0.1& 14.7$\pm$ 0.5 &14.9$\pm$ 0.3 & 13.6$\pm$ 0.4 &14.0$\pm$0.6& 13.1$\pm$ 1.0&18.8$\pm$ 0.0& 45.7$\pm$ 0.3&\textbf{62.3$\pm$ 6.4}\\
		
		&0.2&25.5$\pm$ 0.6 &26.8$\pm$ 0.2 &27.8$\pm$ 0.3 &25.7$\pm$ 0.7& 25.0$\pm$ 0.9 &25.8$\pm$1.2& 25.3$\pm$ 0.8&27.7$\pm$ 0.0& 54.5$\pm$ 0.2&\textbf{73.0$\pm$5.7}\\
		CIFAR100 &0.1&14.5$\pm$ 0.6  &15.0$\pm$ 0.1& 16.2$\pm$ 0.5 &15.0$\pm$ 0.4 & 12.6$\pm$ 0.5& 14.1$\pm$0.4 &14.5$\pm$ 0.3&13.4$\pm$ 0.0& 37.0$\pm$ 0.4&\textbf{61.4$\pm$ 5.5}\\
		
		&0.2&25.6$\pm$ 0.6&27.0$\pm$ 0.2 &27.8$\pm$ 0.6 &26.5$\pm$ 0.4& 24.4$\pm$ 0.9 &25.2$\pm$0.4& 25.8$\pm$ 0.5&23.1$\pm$ 0.0& 49.9$\pm$ 0.3&\textbf{66.3$\pm$ 2.8}\\
		\Xhline{0.5pt} \Xhline{0.5pt}
		\label{Compare}
	\end{tabular}
	\vspace{-1.5mm}
\end{table*}

To demonstrate the superiority of our MCOD, we compare the MCOD with the baselines OC-SVM~\cite{2001Estimating} and nine state-of-the-arts including CAE~\cite{masci2011stacked}, DRAE~\cite{xia2015learning}, DSEBM~\cite{zhai2016deep}, RDAE~\cite{zhou2017anomaly}, MOGAL~\cite{liu2019generative}, RSRAE~\cite{lai2019robust}, RSRAE+~\cite{lai2019robust} SSD-IF~\cite{wangself} and E3Out~\cite{ wangself}. The experiments are conducted on four benchmarks: MNIST~\cite{lecun1995comparison}, Fashion-MNIST~\cite{xiao2017fashion}, CIFAR10~\cite{krizhevsky2009learning}, and CIFAR100~\cite{krizhevsky2009learning}. MCOD is implemented on NVIDIA GTX 2080.
\subsection{Dataset}

\textbf{MNIST}~\cite{lecun1995comparison} contains 70K gray-scale images of ten digits from ``0'' to``9'' size of $28 \times 28$.\\
\indent
\textbf{Fashion-MNIST}(F-MNIST)~\cite{xiao2017fashion} contains 70K clothing and accessories gray-scale images with size of $28 \times 28$.\\
\indent
\textbf{CIFAR10}~\cite{krizhevsky2009learning} contains 60K images with size of $32 \times 32$, which are grouped into ten classes.\\
\indent
\textbf{CIFAR100}~\cite{krizhevsky2009learning} contains the same images in CIFAR10, but has $100$ fine-grained classes or $20$ coarse-grained classes. Following the settings in papers~\cite{wang2019effective, wangself}. we select $20$ coarse-grained classes in our experiment.

\subsection{Evaluation metrics}

To evaluate the outlier detection performance, we adopt the Area under the Receiver Operating Characteristic curve (AUROC) and Area under the Precision-Recall curve (AUPR) as evaluation metrics~\cite{boyd2013area}. The higher AUROC and AUPR represent a better performance. The AUPR scores usually are computed in following ways: taking inliers as positive samples (AUPR-IN) and taking outliers as positive samples (AUPR-OUT). All experiments are performed independently for five times and report the average results.

\subsection{Implementation details}

To test the effectiveness of our MCOD, we follow the standard experimental setting in previous studies~\cite{xia2015learning, zhou2017anomaly, wang2019effective}. Taking the images from the same semantical category as inliers and the random samples from other categories as outliers. We carry out the cross-validation, each category plays as inliers in turn, and report the average performance.

We use Wide ResNet (WRN)~\cite{zagoruyko2016wide} as our backbone network with the widen factor as $4$. The dimensions of encoded features and embedding representations are 256 and 128, respectively. The model is optimized by Adam optimizer with $\beta=(0.9; 0.999)$. The mini-batch size is $256$. The learning rate is $0.0001$ and the weight decay is $0.0005$. For MNIST and F-MNIST, the temperature parameters $\tau _z$ and $\tau _c$ are both set as $1$. For CIFAR10 and CIFAR100, they are set as $10$ and $0.01$, respectively. $\lambda$ is $0.05$ in all experiments. By default, the number of prototypes stored in the memory is $10$. In the training process, we warm-up our network without using memory module in first $100$ epochs, afterwards, include memory in later $100$ epochs.

\begin{figure*}[!htp]
	\centering
	{\includegraphics[width=0.25\linewidth]{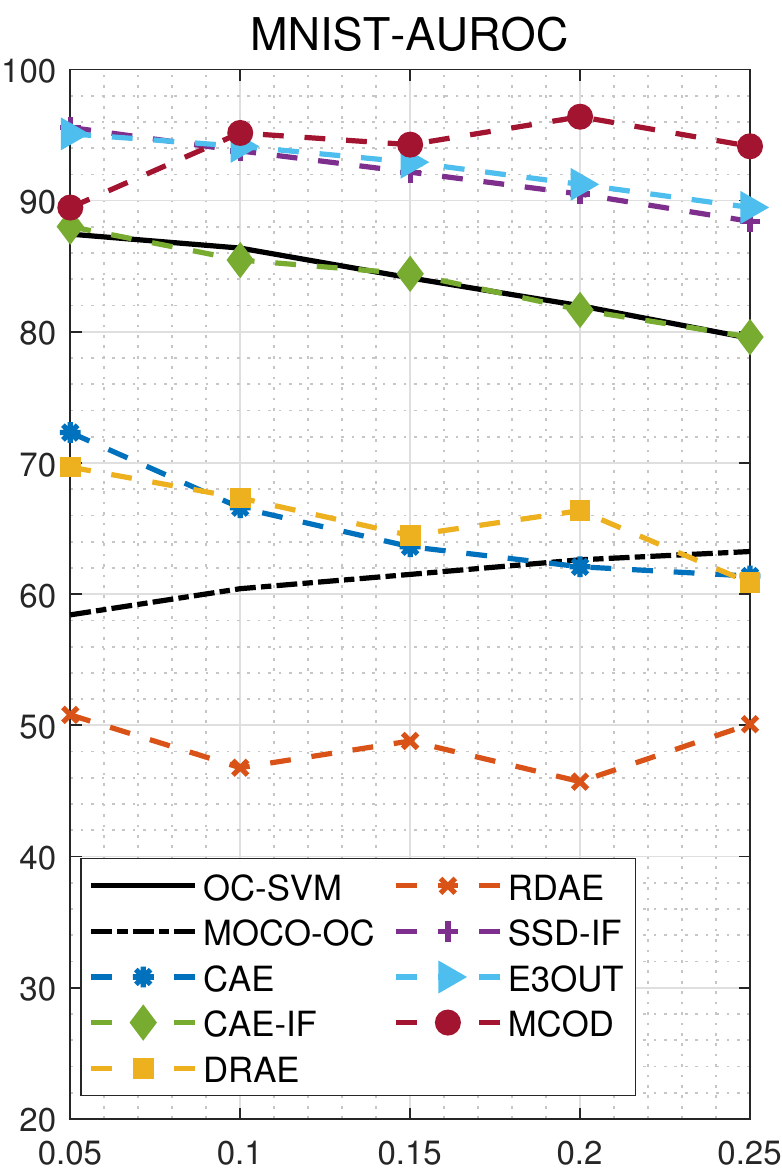}}
	\hspace{-2mm}
	{\includegraphics[width=0.25\linewidth]{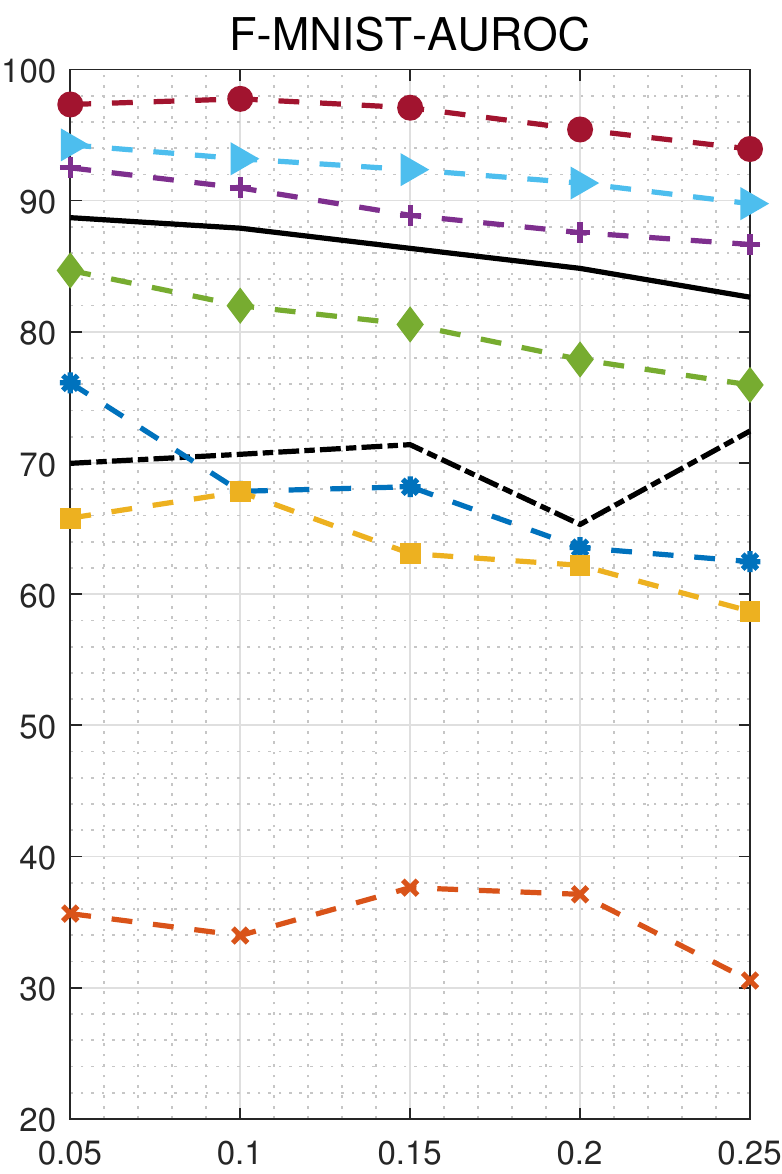}}
	\hspace{-2mm}
	{\includegraphics[width=0.25\linewidth]{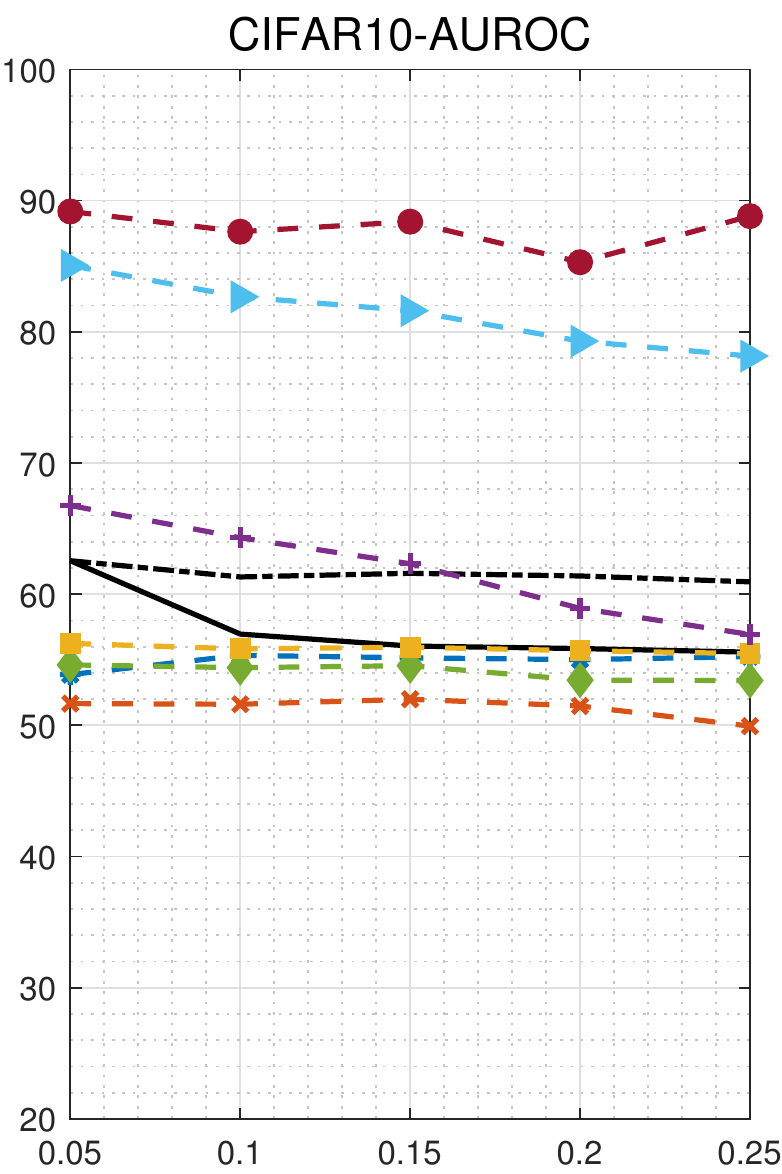}}
	\hspace{-2mm}
	{\includegraphics[width=0.25\linewidth]{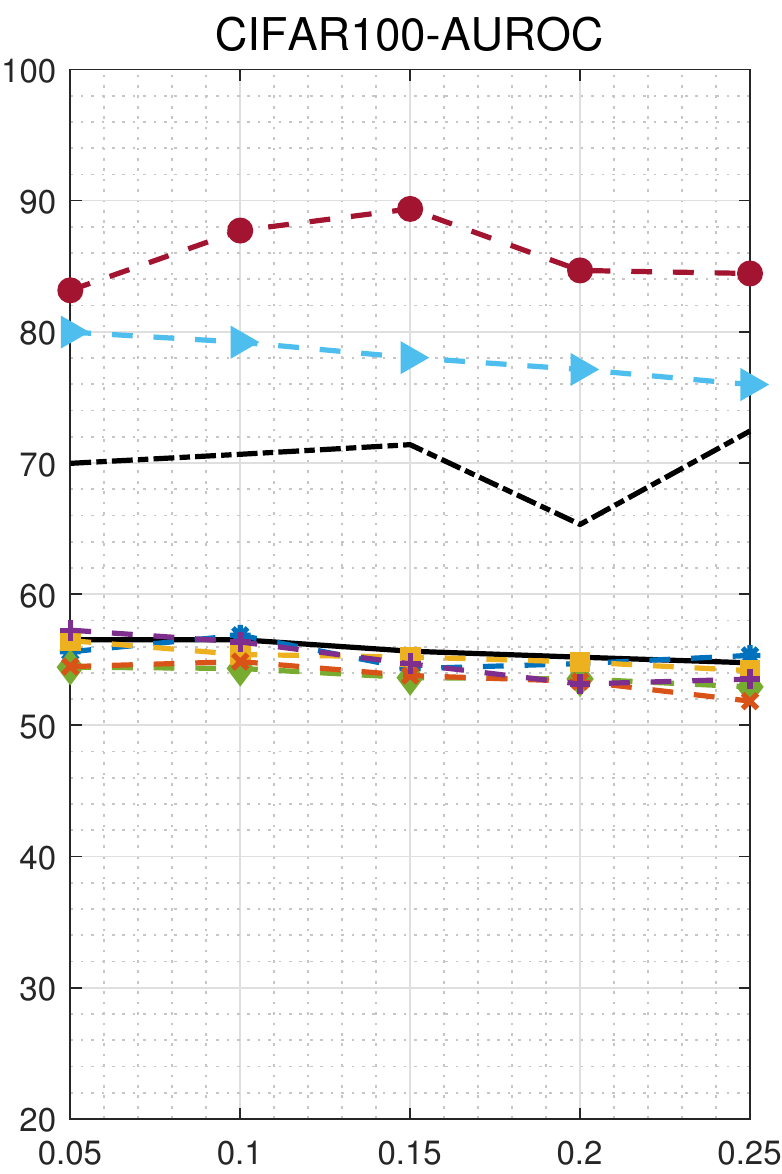}}
	\caption{\footnotesize{Comparison of changes of AUROC ( \% vertical axis) with varied outlier proportion $p$ (horizontal axis) from 0.05 to 0.25. The step is 0.05.}}
	\label{rate}
	\vspace{-1.5mm}
\end{figure*}

\begin{figure*}[!htp]
	\centering
	{\includegraphics[width=0.94\linewidth]{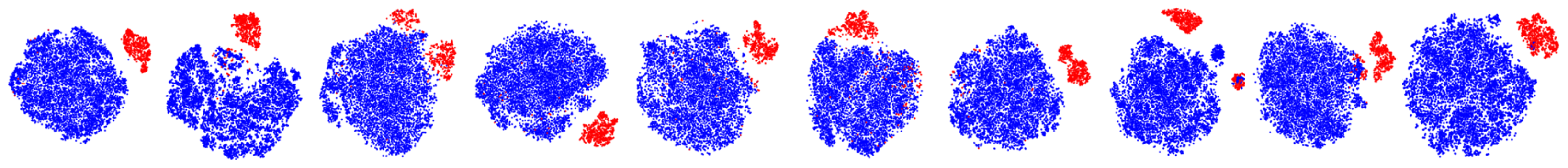}}
	{\includegraphics[width=0.94\linewidth]{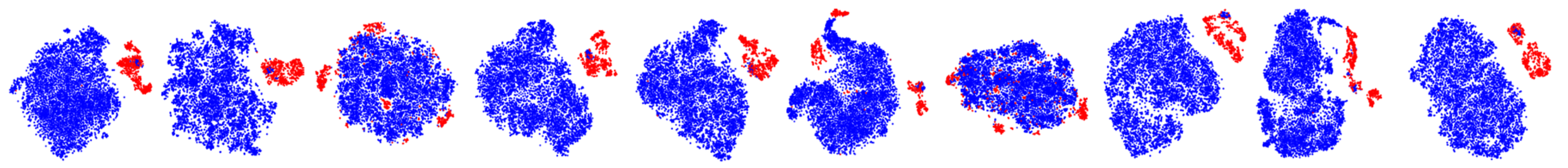}}
	\caption{\footnotesize{T-SNE visualization of learned features of digits $1$ to $10$ in MNIST (first line) and class $1$ to $10$ in F-MNIST (second line). Blue and red dots represent inliers and outliers in the feature space.}}
	\label{TSNE}
\end{figure*}
\subsection{Performance comparison}
We compare the MCOD with the state-of-the-arts including seven AE based methods: CAE~\cite{masci2011stacked}, DRAE~\cite{xia2015learning}, DSEBM~\cite{zhai2016deep}, RDAE~\cite{zhou2017anomaly}, MOGAL~\cite{liu2019generative}, RSRAE~\cite{lai2019robust}, RSRAE+~\cite{lai2019robust} and two self-supervised based methods: SSD-IF~\cite{wangself} and E3Out~\cite{ wangself}. SSD-IF shares the same self-supervised discriminating network with E3Out, but uses the Isolation Forest (IF) \cite{liu2008isolation} as detector. The proportions of outliers in datasets are set to be $p=0.1$ and $p=0.2$. The numerical results of AUROC, AUPR-IN and AUPR-OUT are listed in Table \ref{Compare}. \\
\indent
We can see that our MCOD outperforms most other methods on all datasets in terms of all evaluation metrics. Specially, when $p=0.1$, MCOD achieves a considerable improvement on AUROC (MNIST $28.2\%$, F-MNIST $28.4\%$, CIFAR10 $31.8\%$ and CIFAR100 $36.5\%$) compared with CAE. For the simpler dataset MNIST, MCOD outperforms six AE based methods (DRAE $29.3\%$, DSEBM $35.7\%$, RDAE $24.4\%$, MOGAL $65.3\%$, RSRAE $11.4\%$, RSRAE+ $22.3\%$). For complicated datasets (CIFAR10 and CIFRA100), our superiority become more significant. These results demonstrate that learning consistent and discriminating features does facilitate the performance of outlier detection.\\
\indent
Compared to self-supervised based methods, in terms of the metric AUROC and $p=0.1$, MCOD outperforms SSD-IF (MNIST $2.4\%$, F-MNIST $7.7\%$, CIFAR10 $23.3\%$ and CIFAR100 $31.3\%$) and E3Out (MNIST $1.3\%$, F-MNIST $5.2\%$, CIFAR10 $2.9\%$ and CIFAR100 $6.4\%$), respectively. Although its performs slightly worse than E3Out on MNIST in terms of AUPR-IN, MCOD achieves better scores on F-MNIST, CIFAR10, and CIFAR100. E3Out aims at recognizing the types of sample augmentations, and assumes that the augmentations of inlier samples could be recognized to their corresponding types with a high confidence, but the augmentations of outlier samples could not. Such mechanism of recognizing different types of sample augmentations makes the learned features of outlier and inlier less discriminating. For the simpler dataset MNIST, E3Out achieves a higher AUPR-IN score, but a much lower AUPR-OUT score. For the complicated datasets CIFAR10 and CIFAR100, E3Out commonly performs worse than MCOD in terms of both AUPR-IN and AUPR-OUT scores. This indicates that E3Out is inclined to recognize the simple outliers as inliers and mistakes complicated inlier data as outliers.\\
\indent
Unlike AUPR-IN, AUPR-OUT chooses outliers as positive samples, which is a more intuitive metric of outlier detection. MCOD outperforms all other methods, especially, is better than E3Out with remarkable $14.5\%$ and $12.2\%$ gains on MNIST. When $p=0.2$, the performances of CAE, DSEBM, RDAE, and RSRAE drop dramatically. By contrast, the performance of MCOD is still stable. This significant advantage on AUPR-OUT demonstrates the superiority of our MCOD on learning more discriminating features of outliers. Other methods focus on learning a better fitting model for inlier because inlier is the majority, but consider less on the discrimination of features. By contrast, our MCOD takes both sides into account. The memory model learns the consistent feature to fit inliers. The contrastive learning model forces the features of inlier and outlier more distinct. Therefore, MCOD demonstrates more performance gains on AUPR-OUT against AUPR-IN.\\
\indent
Figure \ref{TSNE} shows the visualization of learned features of inliers and outliers in MNIST and F-MNIST by t-SNE~\cite{maaten2008visualizing}. We can see that outliers are grouped together and keep a distance to inliers in most cases. This intuitive result illustrates the effectiveness of our MCOD. The features of inliers are consistent, meanwhile, the features of outliers are distinct from inliers. It also proves that the outlier detection can be done in the feature space.
\begin{figure}[!tp]
	\centering {\includegraphics[width=0.99\linewidth]{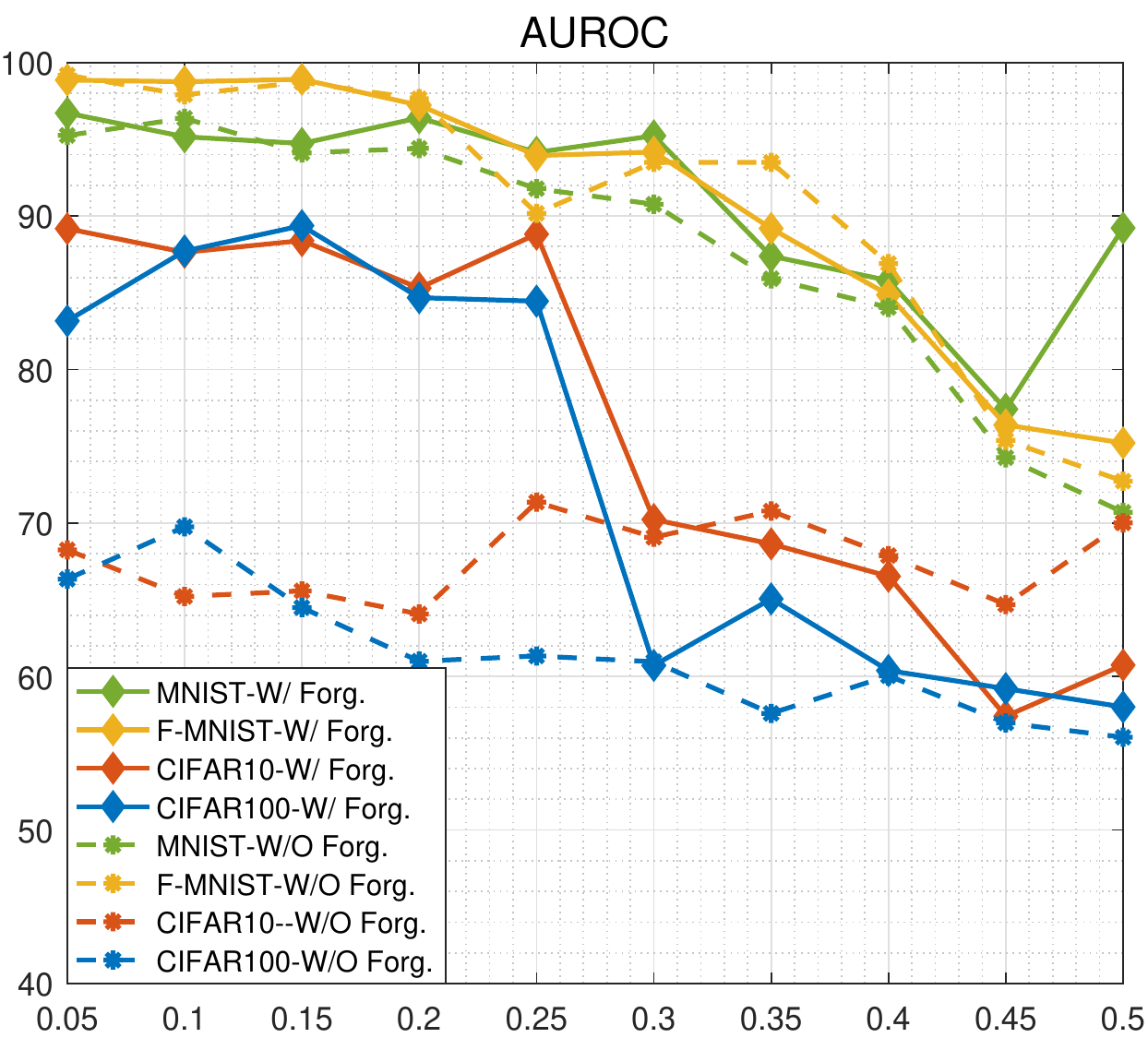}}
	\caption{\footnotesize{AUROC (\% vertical axis) with (W/Forg.) and without forgetting (W/O Forg.) operations with varied outlier proportion $p$ (horizontal axis).}}
	\label{Forgetting}
\end{figure}
\subsection{Varied proportions of outliers}
\begin{figure*}[!htp]
	\centering
	\subfigure[MNIST]
	{\includegraphics[width=0.25\linewidth]{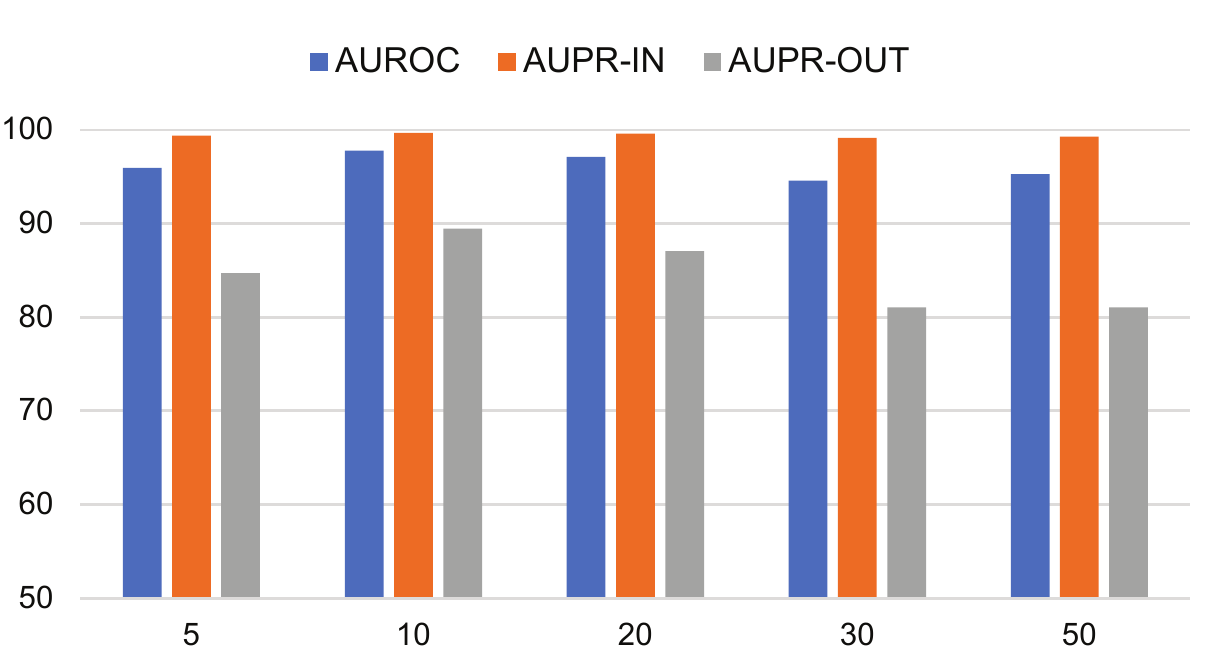}}
	\hspace{-2mm}
	\subfigure[F-MNIST]
	{\includegraphics[width=0.25\linewidth]{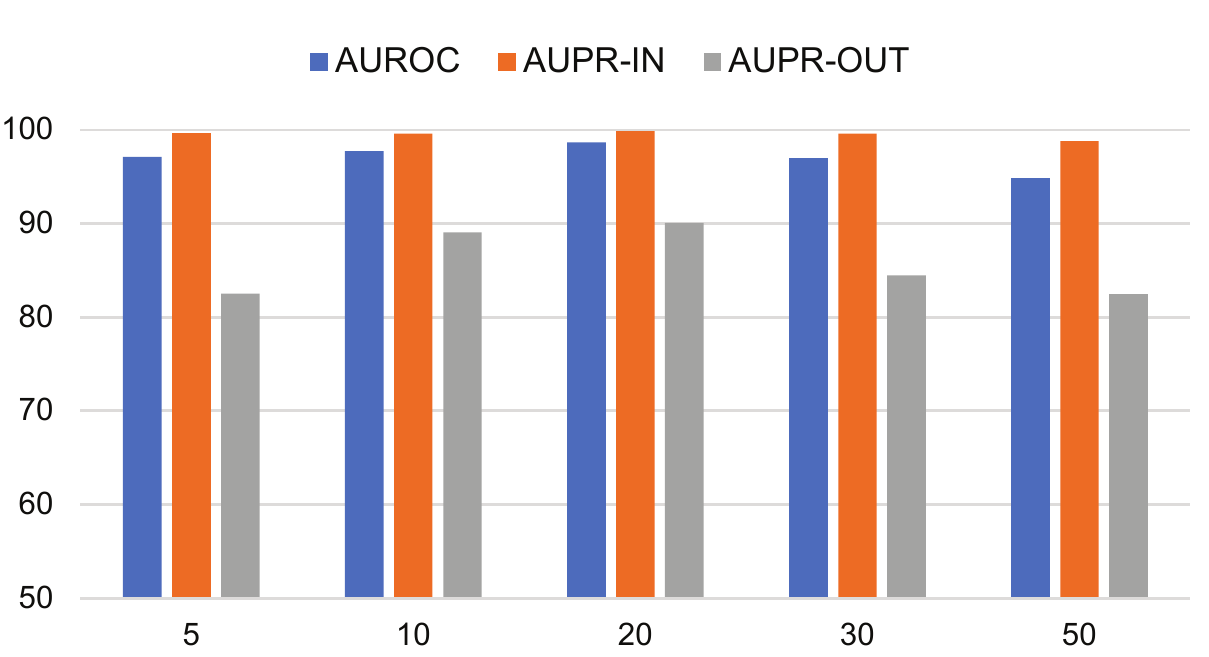}}
	\hspace{-2mm}
	\subfigure[CIFAR10]
	{\includegraphics[width=0.25\linewidth]{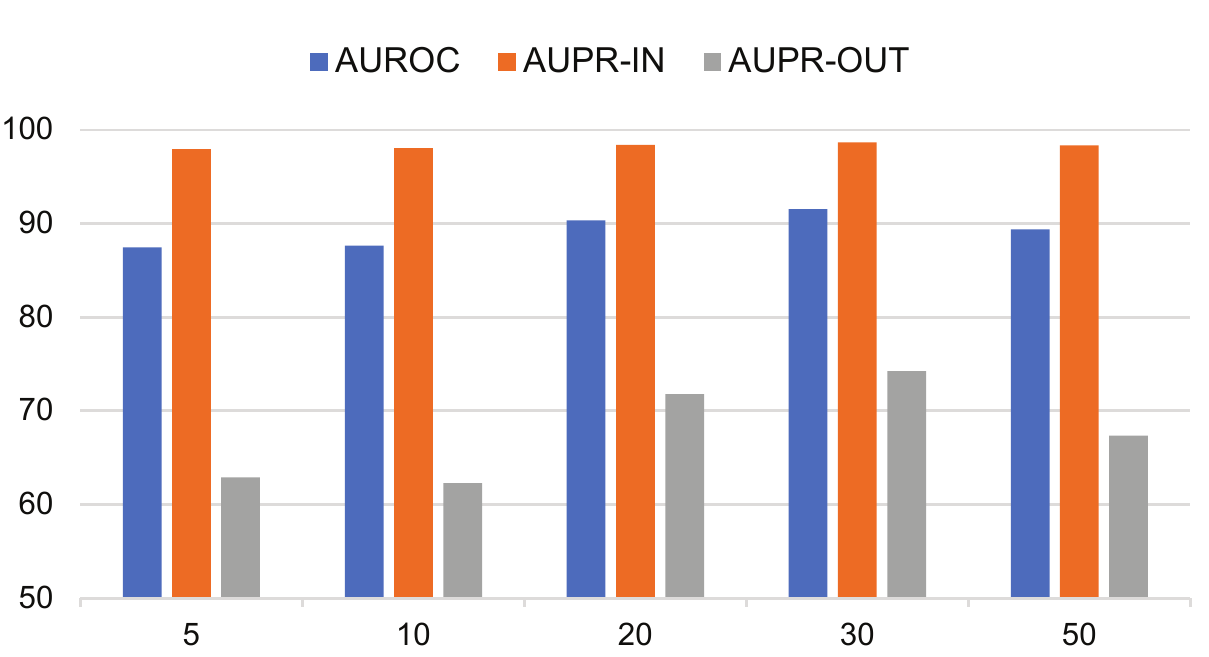}}
	\hspace{-2mm}
    \subfigure[CIFAR100]
	{\includegraphics[width=0.25\linewidth]{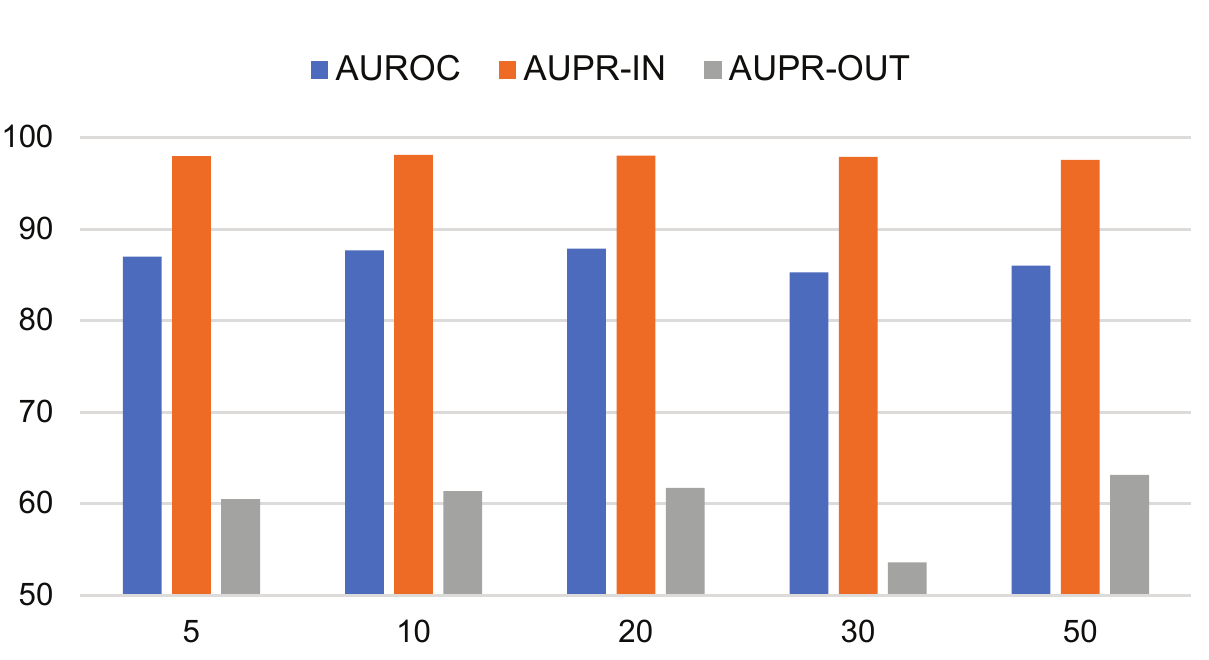}}
	\caption{\footnotesize{AUROC, AUPR-IN, AUPR-OUT (\% vertical axis) with varied number of memory prototypes (horizontal axis) $[5, 10, 20 ,30, 50]$.}}
	\label{Cluster}
\end{figure*}

\begin{figure*}[!htp]
	\centering
	\subfigure[MNIST]
	{\includegraphics[width=0.25\linewidth]{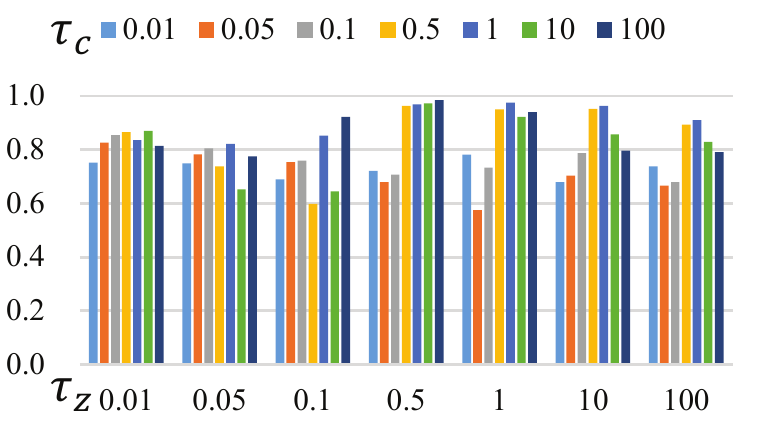}}
	\hspace{-2mm}
	\subfigure[F-MNIST]
	{\includegraphics[width=0.25\linewidth]{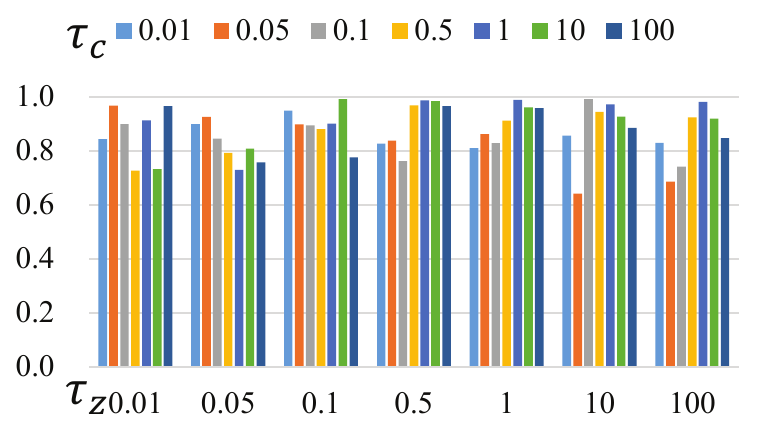}}
	\hspace{-2mm}
	\subfigure[CIFAR10]
	{\includegraphics[width=0.25\linewidth]{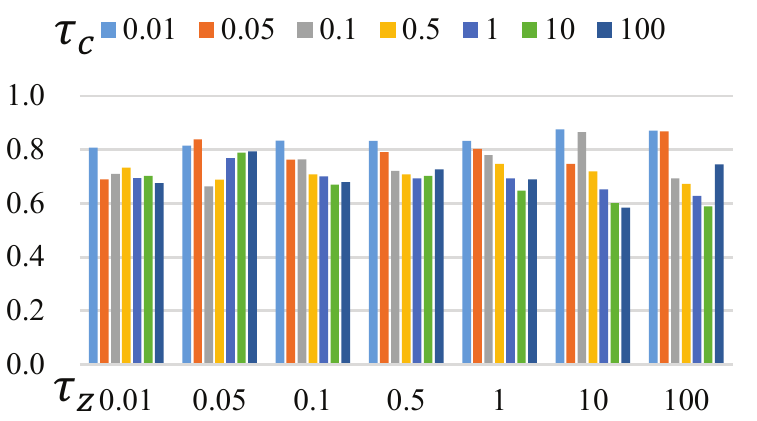}}
	\hspace{-2mm}
	\subfigure[CIFAR100]
	{\includegraphics[width=0.25\linewidth]{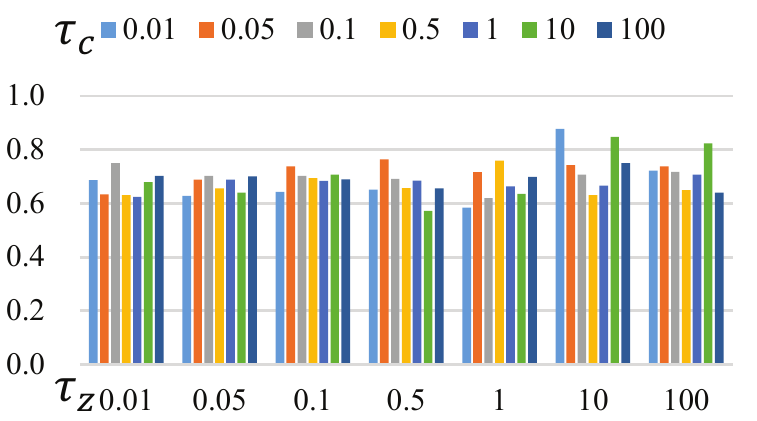}}
	\caption{\footnotesize{AUROC (\% vertical axis) with varied temperature parameters $\tau_z$ (horizontal axis) and $\tau_c$ (different color bins) $[0.01, 0.05, 0.1, 0.5, 1, 10, 100]$.}}
	\label{Temperature}
\end{figure*}

As outlier proportion $p$ can severely affect the performances of OD methods, we test MCOD in terms of AUROC with varied $p=[0.05, 0.1, 0.15, 0.2, 0.25]$. We also compare it with eight competing methods, including two baselines (OC-SVM and MOCO-OC), four AE-based methods (CAE, CAE-IF, DRAE, RDAE), and two self-supervised methods (SSD-IF and E3Out). OC-SVM utilizes the raw data as input. Instead, MOCO-OC uses features learned by MoCo~\cite{he2020momentum} as the input for OC-SVM. CAE-IF uses IF as the outlier detector. The AUROC curves of performance are shown in Fig.~\ref{rate}.

We can see our MCOD demonstrate an obvious superiority on most outlier proportions, except that E3Out and SSD-IF show a better AUROC score only when $p = 0.05$. With the outlier proportion increasing, the gap between MCOD and others is gradually enlarged. On MNIST and F-MNIST, the comparable methods are E3Out and SSD-IF. On complicated datasets CIFAR10 and CIFAR100, only E3Out is comparable with MCOD. This result indicate that self-supervised methods is superior than AE based methods. And the contrastive learning can contribute more performance gain. The possible reason could be following. As $p$ increases, outliers (randomly sampled data) are more likely to join to AE model training. Thus, AE based methods result in a better outlier reconstruction. For the self-supervised method, E3Out is based on the accuracy of recognizing the types of sample augmentations. With more outliers being involved, E3Out may predict the type of outlier augmentation with a less uncertainty (higher confidence), which degrade the performance of outlier detection. Although our MCOD also faces more outliers, the memory model constrains the consistency of inlier features. It makes MCOD detect outlier easier by measuring the distance between samples and their corresponding prototypes.\\
\subsection{Ablation study}

\textbf{Evaluation of memory forgetting operation.} Since the forgetting operation plays a crucial role of suppressing the outlier prototypes in memory slots, we report the AUROC scores with (W/ Forg.) and without forgetting operations (W/O Forg.) when vary $p$ from $0.05$ to $0.5$ in Fig.~\ref{Forgetting}. We can see MCOD (W/ Forg.) shows a significant superiority than MCOD (W/O Forg.) on more complicated datasets CIFAR10 and CIFAR100 when $p<0.3$. But MCOD (W/ Forg.) quickly downgrades when $p \ge 0.3$, because it breaks the assumption that outliers are minority. For the simpler datasets MNIST and F-MNIST, we can observe a similar trend. Since images in MNIST and F-MNIST are simpler, the memory model is easy to learn the consistent features of inliers. Thus, our proposed forgetting operator demonstrates less advantages. Images in CIFAR10 and CIFAR100 are complex and diverse. Such random outliers in the data disrupt the consistent features learning. However, our proposed forgetting operation can alleviate this influence, and demonstrates an obvious performance gain on CIFAR10 and CIFAR100 when $p < 0.3$.

\textbf{Evaluation on varied memory prototype numbers.} Since the memory prototypes is essential for learning the consistent feature of inlier, we report performance of MCOD with varied number of memory prototypes $[5, 10, 20, 30 ,50]$ in terms of AUROC, AUPR-IN, and AUPR-OUT on MNIST, F-MNIST, CIFAR10, and CIFAR100 dataset. The experiments is conducted with outliers proportion $p=0.1$. Figure~\ref{Cluster} illustrates that MCOD is insensitive to the change of prototype numbers. Considering both performances and efficiency, we select the prototypes number as 10.

\textbf{Evaluation on varied temperature parameters.} We report the AUROC scores with different temperature parameters $\tau_z$ and $\tau_c$ in Eq. \ref{feature_infoNCE} and \ref{cluster_infoNCE}. Each of them is set as $[0.01, 0.05, 0.5, 1, 10, 100]$ and the results are reported in Fig. \ref{Temperature}. According to this result, we set $\tau_z = 1$ and $\tau_c = 1$ for MNIST and F-MNIST, and $\tau_z = 10$ and $\tau_c = 0.01$ for CIFAR10 and CIFAR100.
\section{Conclusion}
In this paper, we proposed an unsupervised outlier detection framework using a memory module with specially designed reading, writing, and forgetting operations and a contrastive learning module. Unlike conventional AE based methods detecting outliers according to the reconstruction errors in the pixel space, our proposed MCOD distinguishes outliers and inliers in the feature space. The experimental results on four benchmark datasets, which have various data distributions, have demonstrated the superiority of our MCOD compared to nine stat-of-the-art methods. The detailed analysis shows that the memory module does learn consistent features for inliers, and the contrastive learning module extracts the discriminating features to separate outliers from inliers. Our future work will investigate prior features of inliers to strengthen the feature consistency.

{\small
\bibliographystyle{ieee_fullname}
\bibliography{egpaper}

\begin{thebibliography}{10}\itemsep=-1pt

\bibitem{bourlard1988auto}
Herv{\'e} Bourlard and Yves Kamp.
\newblock Auto-association by multilayer perceptrons and singular value
  decomposition.
\newblock {\em Biological cybernetics}, 59(4-5):291--294, 1988.

\bibitem{boyd2013area}
Kendrick Boyd, Kevin~H Eng, and C~David Page.
\newblock Area under the precision-recall curve: point estimates and confidence
  intervals.
\newblock In {\em ECML-PKDD}, pages 451--466. Springer, 2013.

\bibitem{Chalapathy2019Deep}
Raghavendra Chalapathy and Sanjay Chawla.
\newblock Deep learning for anomaly detection: A survey.
\newblock {\em preprint arXiv:1901.03407}, 2019.

\bibitem{chen2020improved}
Xinlei Chen, Haoqi Fan, Ross Girshick, and Kaiming He.
\newblock Improved baselines with momentum contrastive learning.
\newblock {\em preprint arXiv:2003.04297}, 2020.

\bibitem{chopra2005learning}
Sumit Chopra, Raia Hadsell, and Yann LeCun.
\newblock Learning a similarity metric discriminatively, with application to
  face verification.
\newblock In {\em CVPR}, volume~1, pages 539--546, 2005.

\bibitem{eduardo2020robust}
Simao Eduardo, Alfredo Naz{\'a}bal, Christopher~KI Williams, and Charles
  Sutton.
\newblock Robust variational autoencoders for outlier detection and repair of
  mixed-type data.
\newblock In {\em AISTATS}, pages 4056--4066, 2020.

\bibitem{gong2019memorizing}
Dong Gong, Lingqiao Liu, Vuong Le, Budhaditya Saha, Moussa~Reda Mansour, Svetha
  Venkatesh, and Anton van~den Hengel.
\newblock Memorizing normality to detect anomaly: Memory-augmented deep
  autoencoder for unsupervised anomaly detection.
\newblock In {\em ICCV}, pages 1705--1714, 2019.

\bibitem{graves2014neural}
Alex Graves, Greg Wayne, and Ivo Danihelka.
\newblock Neural turing machines.
\newblock {\em preprint arXiv:1410.5401}, 2014.

\bibitem{he2020momentum}
Kaiming He, Haoqi Fan, Yuxin Wu, Saining Xie, and Ross Girshick.
\newblock Momentum contrast for unsupervised visual representation learning.
\newblock In {\em CVPR}, pages 9729--9738, 2020.

\bibitem{jing2020self}
Longlong Jing and Yingli Tian.
\newblock Self-supervised visual feature learning with deep neural networks: A
  survey.
\newblock {\em IEEE TPAMI}, 2020.

\bibitem{krizhevsky2009learning}
Alex Krizhevsky, Geoffrey Hinton, et~al.
\newblock Learning multiple layers of features from tiny images.
\newblock 2009.

\bibitem{lai2019robust}
Chieh-Hsin Lai, Dongmian Zou, and Gilad Lerman.
\newblock Robust subspace recovery layer for unsupervised anomaly detection.
\newblock {\em preprint arXiv:1904.00152}, 2019.

\bibitem{lecun1995comparison}
Yann LeCun, LD Jackel, Leon Bottou, A Brunot, Corinna Cortes, J Denker, Harris
  Drucker, I Guyon, UA Muller, Eduard Sackinger, et~al.
\newblock Comparison of learning algorithms for handwritten digit recognition.
\newblock In {\em ICANN}.

\bibitem{lee2018memory}
Sangho Lee, Jinyoung Sung, Youngjae Yu, and Gunhee Kim.
\newblock A memory network approach for story-based temporal summarization of
  360 videos.
\newblock In {\em CVPR}, pages 1410--1419, 2018.

\bibitem{liang2017enhancing}
Shiyu Liang, Yixuan Li, and Rayadurgam Srikant.
\newblock Enhancing the reliability of out-of-distribution image detection in
  neural networks.
\newblock {\em preprint arXiv:1706.02690}, 2017.

\bibitem{liu2008isolation}
Fei~Tony Liu, Kai~Ming Ting, and Zhi-Hua Zhou.
\newblock Isolation forest.
\newblock In {\em ICDM}, pages 413--422, 2008.

\bibitem{liu2019generative}
Yezheng Liu, Zhe Li, Chong Zhou, Yuanchun Jiang, Jianshan Sun, Meng Wang, and
  Xiangnan He.
\newblock Generative adversarial active learning for unsupervised outlier
  detection.
\newblock {\em IEEE TKDE}, 2019.

\bibitem{maaten2008visualizing}
Laurens van~der Maaten and Geoffrey Hinton.
\newblock Visualizing data using t-sne.
\newblock {\em JMLR}, 9(Nov):2579--2605, 2008.

\bibitem{masci2011stacked}
Jonathan Masci, Ueli Meier, Dan Cire{\c{s}}an, and J{\"u}rgen Schmidhuber.
\newblock Stacked convolutional auto-encoders for hierarchical feature
  extraction.
\newblock In {\em ICANN}.

\bibitem{oord2018representation}
Aaron van~den Oord, Yazhe Li, and Oriol Vinyals.
\newblock Representation learning with contrastive predictive coding.
\newblock {\em preprint arXiv:1807.03748}, 2018.

\bibitem{park2020learning}
Hyunjong Park, Jongyoun Noh, and Bumsub Ham.
\newblock Learning memory-guided normality for anomaly detection.
\newblock In {\em CVPR}, pages 14372--14381, 2020.

\bibitem{perera2019ocgan}
Pramuditha Perera, Ramesh Nallapati, and Bing Xiang.
\newblock Ocgan: One-class novelty detection using gans with constrained latent
  representations.
\newblock In {\em CVPR}, pages 2898--2906, 2019.

\bibitem{pmlr-v80-ruff18a}
Lukas Ruff, Robert~A. Vandermeulen, Nico G{\"o}rnitz, Lucas Deecke, Shoaib~A.
  Siddiqui, Alexander Binder, Emmanuel M{\"u}ller, and Marius Kloft.
\newblock Deep one-class classification.
\newblock In {\em ICML}, volume~80, pages 4393--4402, 2018.

\bibitem{sabokrou2018adversarially}
Mohammad Sabokrou, Mohammad Khalooei, Mahmood Fathy, and Ehsan Adeli.
\newblock Adversarially learned one-class classifier for novelty detection.
\newblock In {\em CVPR}, pages 3379--3388, 2018.

\bibitem{santoro2016one}
Adam Santoro, Sergey Bartunov, Matthew Botvinick, Daan Wierstra, and Timothy
  Lillicrap.
\newblock One-shot learning with memory-augmented neural networks.
\newblock {\em preprint arXiv:1605.06065}, 2016.

\bibitem{schlegl2017unsupervised}
Thomas Schlegl, Philipp Seeb{\"o}ck, Sebastian~M Waldstein, Ursula
  Schmidt-Erfurth, and Georg Langs.
\newblock Unsupervised anomaly detection with generative adversarial networks
  to guide marker discovery.
\newblock In {\em IPMI}.

\bibitem{2001Estimating}
Bernhard Sch{\"o}Lkopf, John~C. Platt, John Shawe-Taylor, Alex~J. Smola, and
  Robert~C. Williamson.
\newblock Estimating the support of a high-dimensional distribution.
\newblock {\em Neural Computation}, 13(7):1443--1471, 2001.

\bibitem{scholkopf1999support}
Bernhard Sch{\"o}lkopf, Robert~C Williamson, Alexander~J Smola, John
  Shawe-Taylor, John~C Platt, et~al.
\newblock Support vector method for novelty detection.
\newblock In {\em NIPS}, volume~12, pages 582--588, 1999.

\bibitem{TaxSVDD}
David M.~J. Tax and Robert P.~W. Duin.
\newblock Support vector data description.
\newblock {\em Mach. Learn.}, 54(1):45--66, Jan 2004.

\bibitem{wangself}
Siqi Wang, Yijie Zeng, Xinwang Liu, Sihang Zhou, En Zhu, Marius Kloft, Jianping
  Yin, and Qing Liao.
\newblock Self-supervised deep outlier removal with network uncertainty and
  score refinement.
\newblock 2020.

\bibitem{wang2019effective}
Siqi Wang, Yijie Zeng, Xinwang Liu, En Zhu, Jianping Yin, Chuanfu Xu, and
  Marius Kloft.
\newblock Effective end-to-end unsupervised outlier detection via inlier
  priority of discriminative network.
\newblock In {\em NIPS}, pages 5962--5975, 2019.

\bibitem{wang2020cluster}
Ziming Wang, Yuexian Zou, and Zeming Zhang.
\newblock Cluster attention contrast for video anomaly detection.
\newblock In {\em ACMM}, pages 2463--2471, 2020.

\bibitem{xia2015learning}
Yan Xia, Xudong Cao, Fang Wen, Gang Hua, and Jian Sun.
\newblock Learning discriminative reconstructions for unsupervised outlier
  removal.
\newblock In {\em ICCV}, pages 1511--1519, 2015.

\bibitem{xiao2017fashion}
Han Xiao, Kashif Rasul, and Roland Vollgraf.
\newblock Fashion-mnist: a novel image dataset for benchmarking machine
  learning algorithms.
\newblock {\em preprint arXiv:1708.07747}, 2017.

\bibitem{zagoruyko2016wide}
Sergey Zagoruyko and Nikos Komodakis.
\newblock Wide residual networks.
\newblock {\em preprint arXiv:1605.07146}, 2016.

\bibitem{zhai2016deep}
Shuangfei Zhai, Yu Cheng, Weining Lu, and Zhongfei Zhang.
\newblock Deep structured energy based models for anomaly detection.
\newblock {\em preprint arXiv:1605.07717}, 2016.

\bibitem{zhong2020deep}
Huasong Zhong, Chong Chen, Zhongming Jin, and Xian-Sheng Hua.
\newblock Deep robust clustering by contrastive learning.
\newblock {\em preprint arXiv:2008.03030}, 2020.

\bibitem{zhou2017anomaly}
Chong Zhou and Randy~C Paffenroth.
\newblock Anomaly detection with robust deep autoencoders.
\newblock In {\em ACM SIGKDD}, pages 665--674, 2017.

\bibitem{2018A}
Xun Zhou, Sicong Cheng, Meng Zhu, Chengkun Guo, Sida Zhou, Peng Xu, Zhenghua
  Xue, and Weishi Zhang.
\newblock A state of the art survey of data mining-based fraud detection and
  credit scoring.
\newblock {\em MATEC Web of Conferences}, 189:03002, 01 2018.

\bibitem{zhu2019dm}
Minfeng Zhu, Pingbo Pan, Wei Chen, and Yi Yang.
\newblock Dm-gan: Dynamic memory generative adversarial networks for
  text-to-image synthesis.
\newblock In {\em CVPR}, pages 5802--5810, 2019.

\bibitem{zong2018deep}
Bo Zong, Qi Song, Martin~Renqiang Min, Wei Cheng, Cristian Lumezanu, Daeki Cho,
  and Haifeng Chen.
\newblock Deep autoencoding gaussian mixture model for unsupervised anomaly
  detection.
\newblock In {\em ICLR}, 2018.

\end{thebibliography}
}
\clearpage
\begin{LARGE}
\noindent{\textbf{Supplementary}}
\end{LARGE}
\appendix
\section{Evaluation on more datasets}

To evaluate the robustness of MCOD, we test it on two more challenging datasets: STL-10 and Tiny Imagenet. They both have less training data compared to the four datasets in main manuscript. Moreover, the inliers and outliers in Tiny Imagenet are much more divers, which raises the difficulty of outlier detection.

\textbf{STL-10} contains $500$ training images, $800$ test images per class (10 classes) with size of $96 \times 96$. In our experiment, we combine training and test images together to get $1.3K$ images per class. The images from the same semantical category are taken as inliers and the random samples from other categories are considered as outliers.

\textbf{Tiny Imagenet} contains $200$ classes of images from a distinct subset of Imagenet with size of $64 \times 64$. We select the first $20$ categories with $500$ images per category as inliers respectively. The outliers are from other $199$ categories besides of the current inliers category.

The data setting is as the same as that in the main manuscript. The temperature parameters $\tau_z$ and $\tau_c$ are set as $10$ and $0.01$ for both STL-10 and Tiny Imagenet. Other implementation details are set as the same as in Section 4.3 of the main manuscript. We run each scenario $5$ times and report the average results in table \ref{table_result}, when $p=0.1$ and $p=0.2$. MCOD demonstrates a considerable effectiveness on both datasets. In terms of AUROC, MCOD achieves $94.8\%$ ($p=0.1$) and $89.0\%$ ($p=0.2$) on STL-10. For more complicated outliers MCOD gets $81.8\%$ ($p=0.1$) and $79.2\%$ ($p=0.2$) and $81.8\%$ on Tiny Imagenet. When the proportion of outliers increases, the performance of MCOD on these two datasets decreases slightly ($89.0\%$ on STL-10 and $79.2\%$ on Tiny Imagenet) which shows the consistent tendency as discussed in the main manuscript.

\section{The visualization of feature distribution of MNIST and F-MNIST}

\begin{table}[!h]
	\footnotesize
	\centering
	\caption {\footnotesize{Results of MCOD on STL-10 and Tiny Imagenet}}
	\begin{tabular}{cccccc}
		\hline
		Dataset    & $p$   & AUROC\%    & AUPR-IN\%    & AUPR-OUT\% \\ \hline
		STL-10    & 0.1 & 94.8 & 99.3 & 70.8 \\
		& 0.2 & 89.0 & 96.4 & 68.8 \\ \hline
		Tiny Imagenet  & 0.1 & 81.8 & 97.0 & 48.8\\
		& 0.2 & 79.7 & 92.6 & 59.3\\ \hline
		
	\end{tabular}
	\label{table_result}
\end{table}

\begin{figure*}[!htp]
	\centering
	\subfigure[``0'']
	{\includegraphics[width=0.3\linewidth]{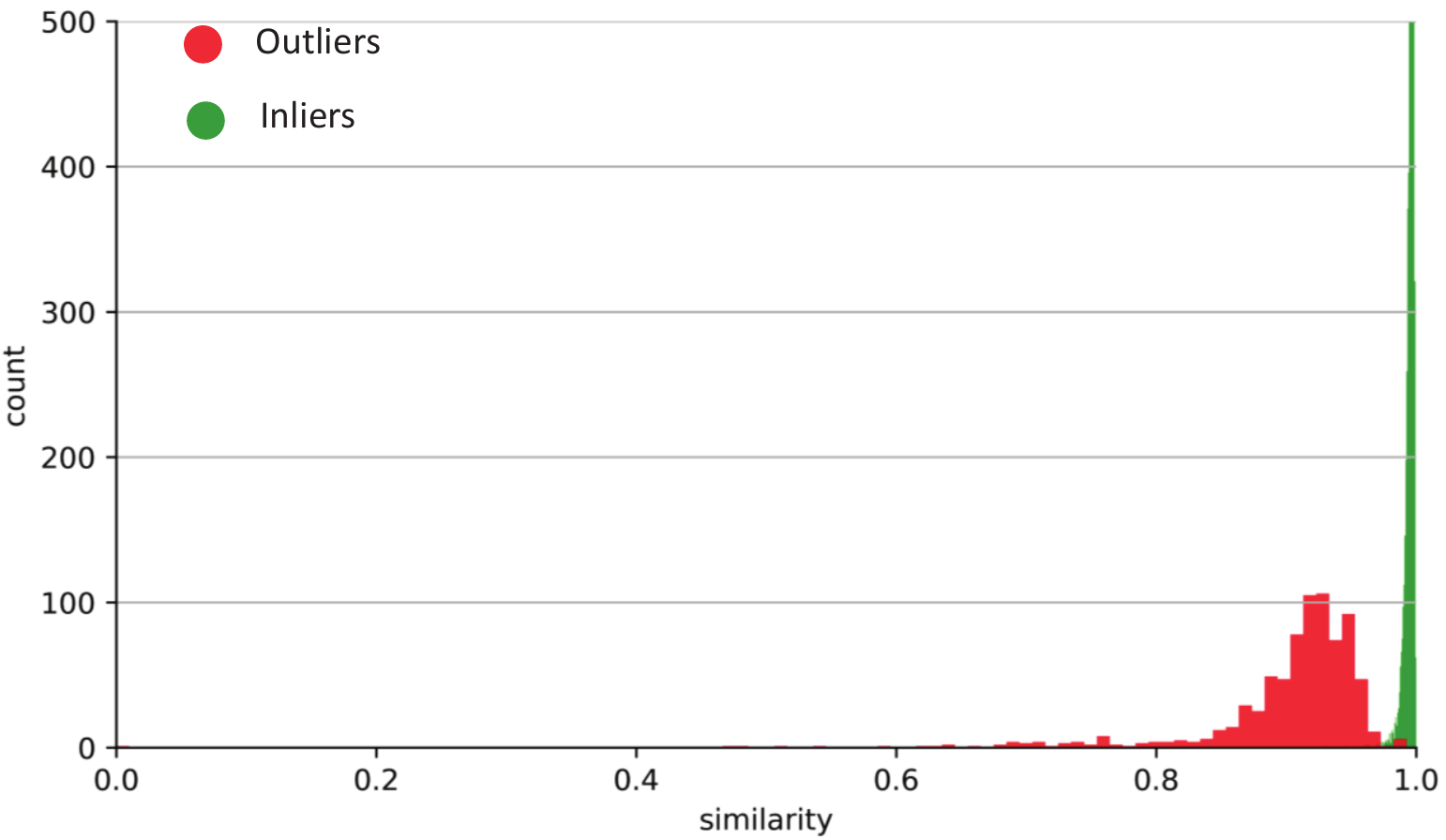}}
	\subfigure[``1'']
	{\includegraphics[width=0.3\linewidth]{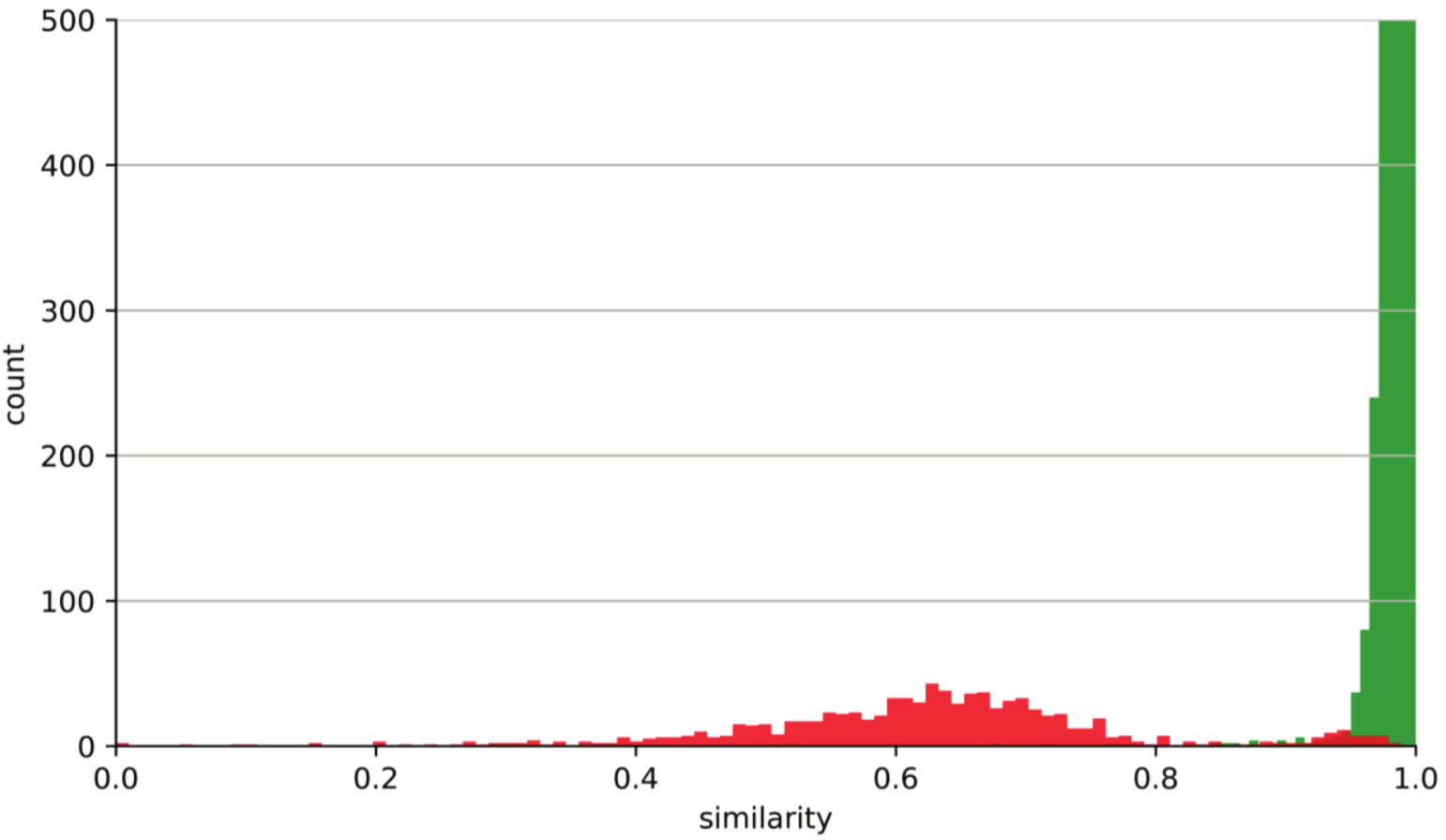}}
	\subfigure[``2'']
	{\includegraphics[width=0.3\linewidth]{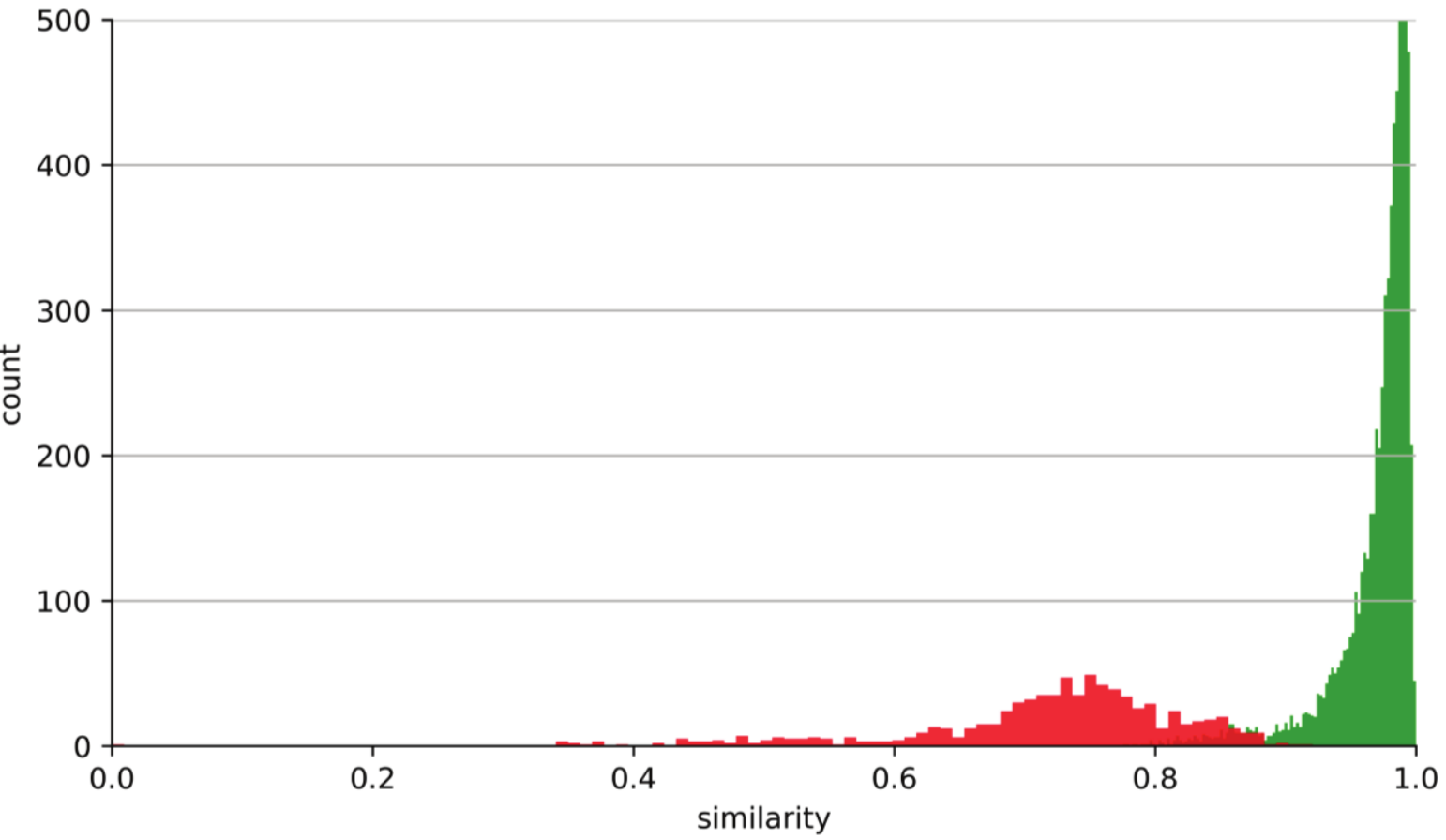}}
	\subfigure[``3'']
	{\includegraphics[width=0.3\linewidth]{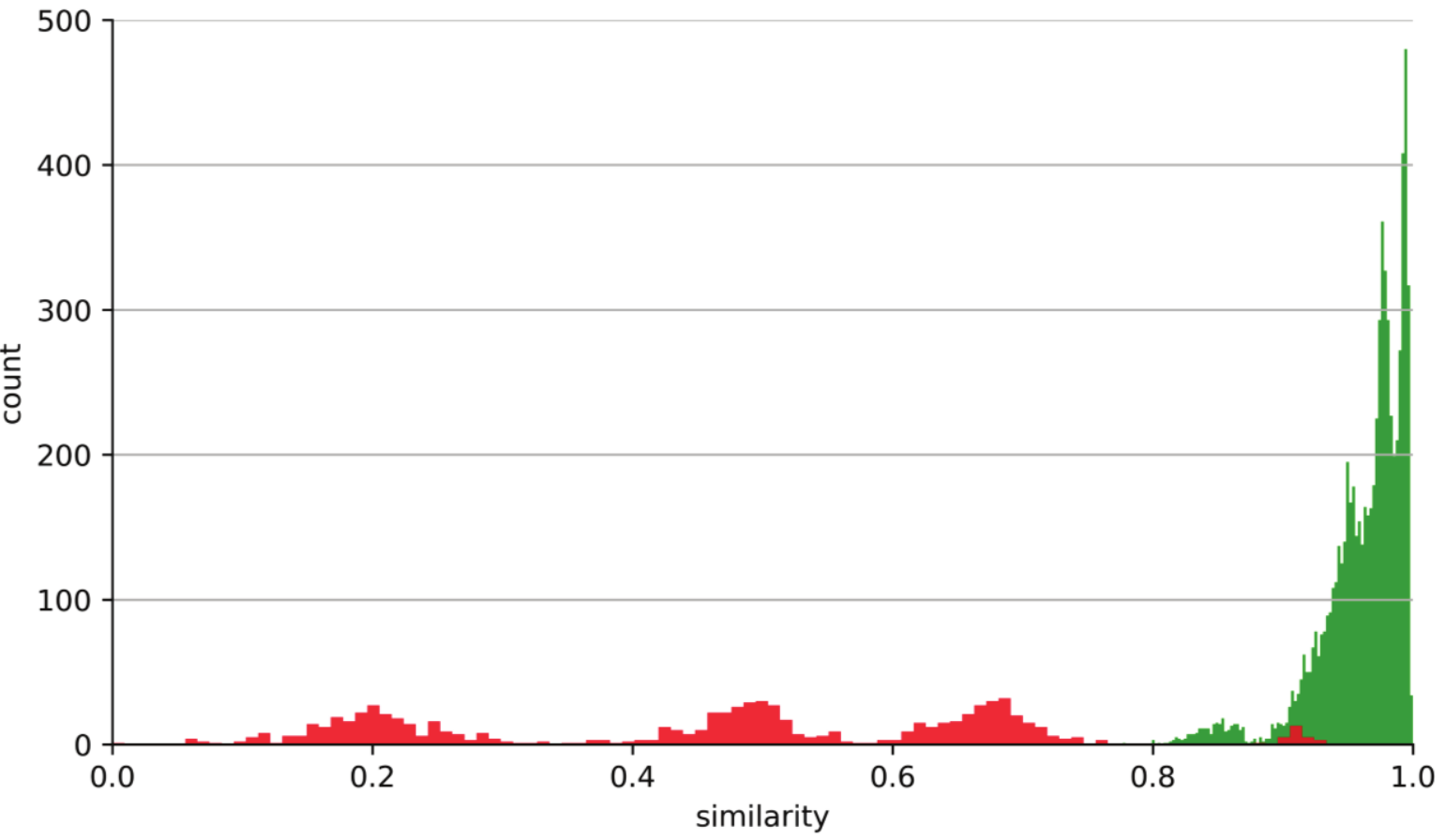}}
	\subfigure[``4'']
	{\includegraphics[width=0.3\linewidth]{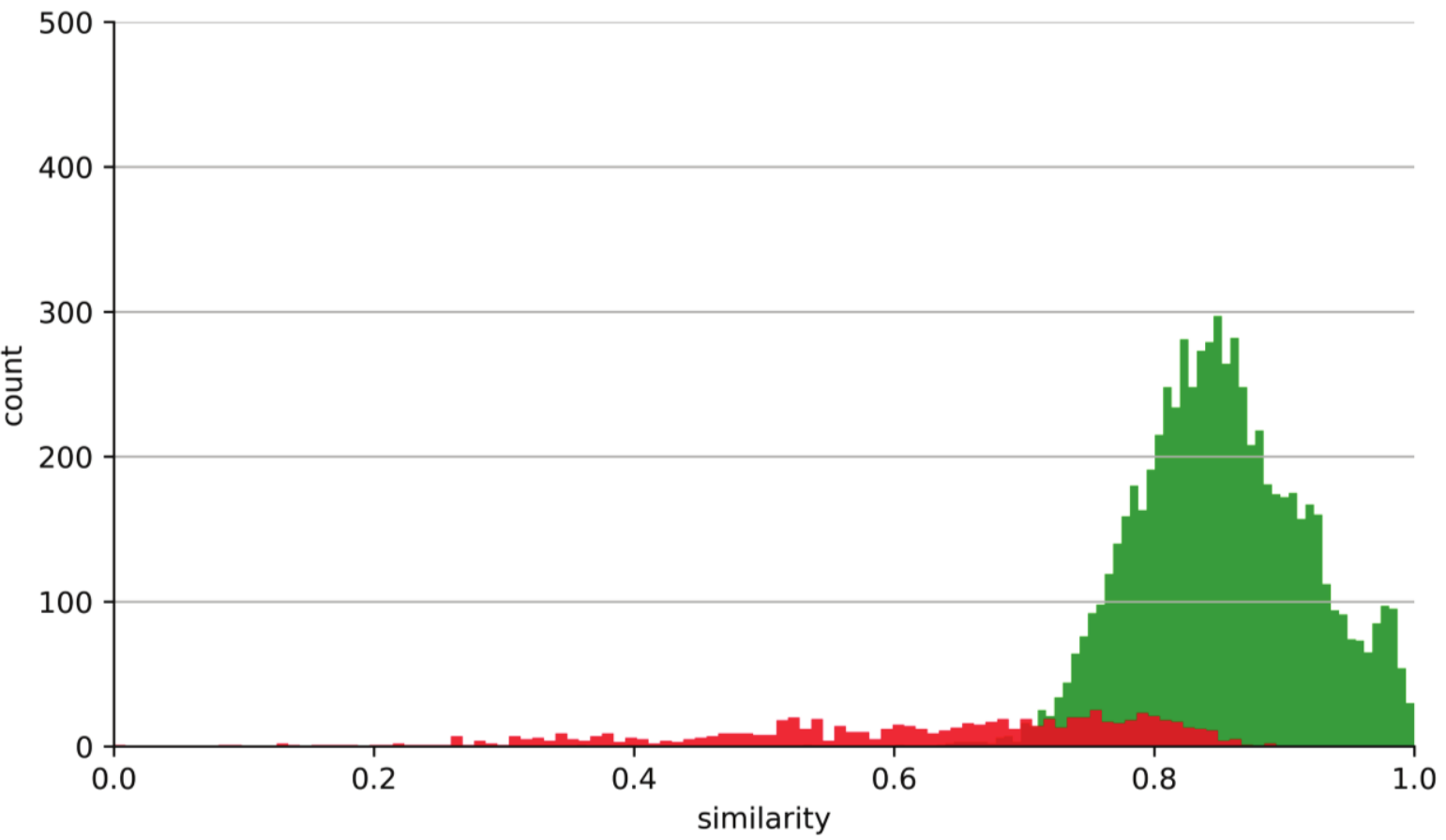}}
	\subfigure[``5'']
	{\includegraphics[width=0.3\linewidth]{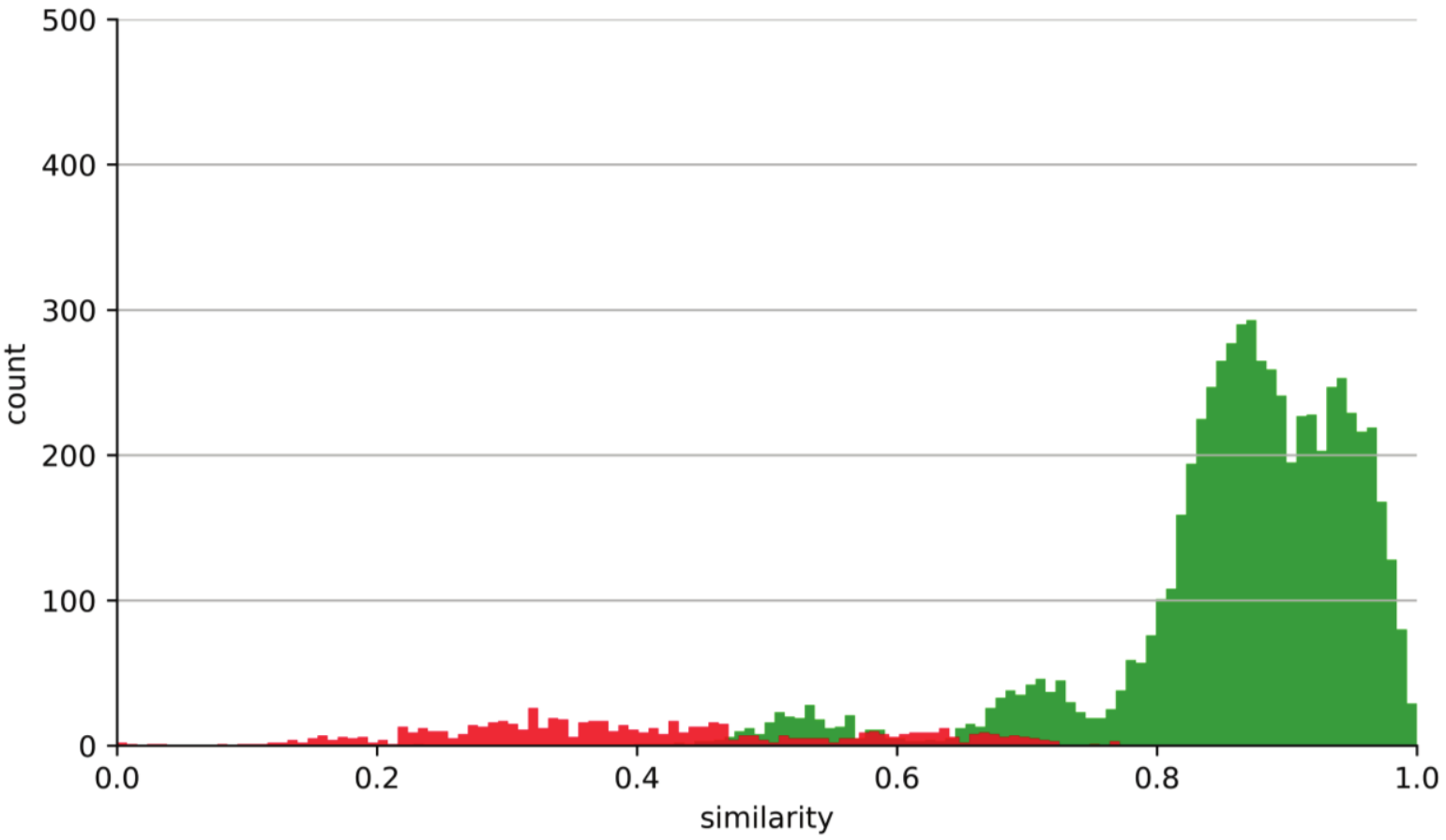}}
	\subfigure[``6'']
	{\includegraphics[width=0.3\linewidth]{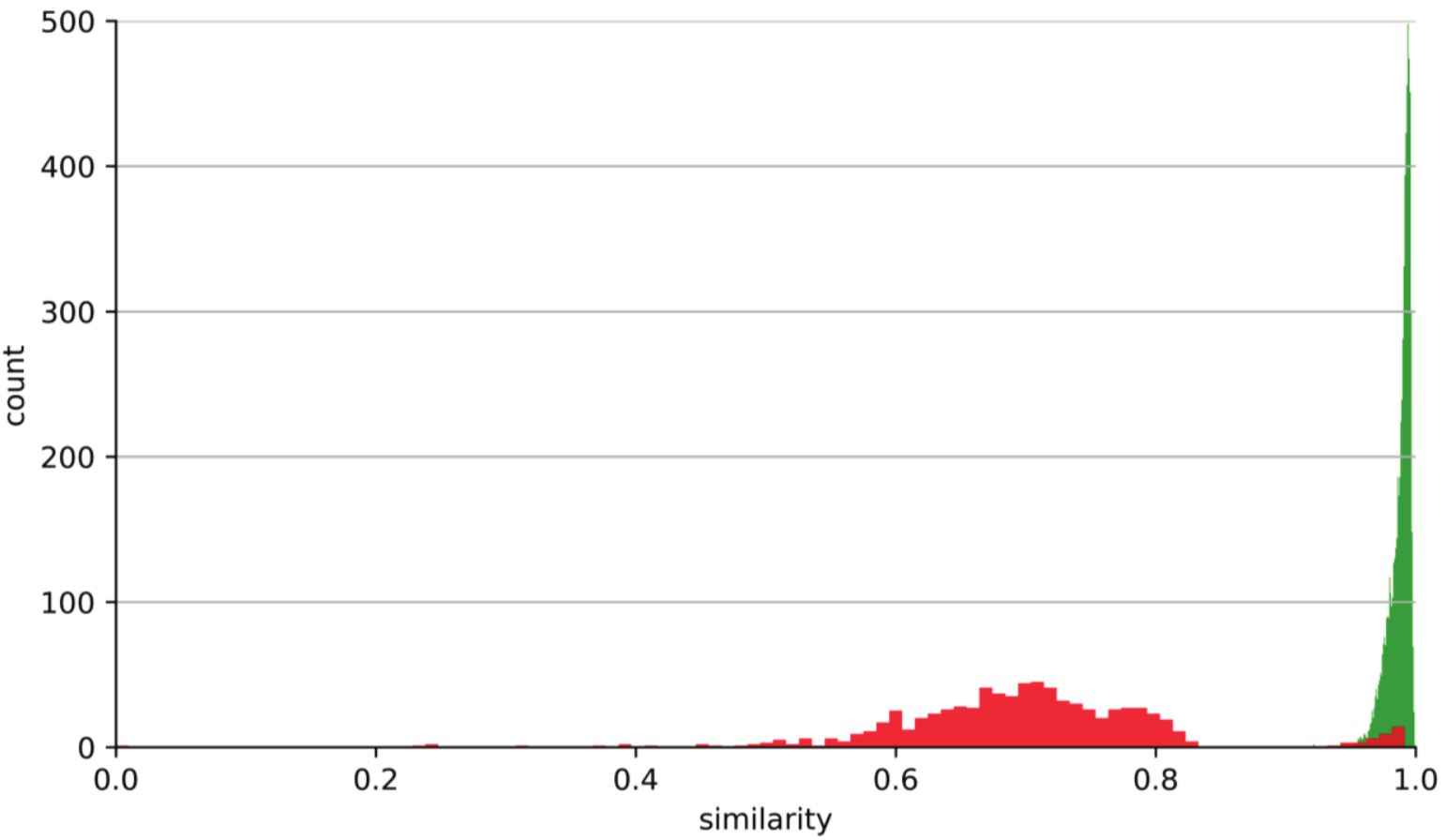}}
	\subfigure[``7'']
	{\includegraphics[width=0.3\linewidth]{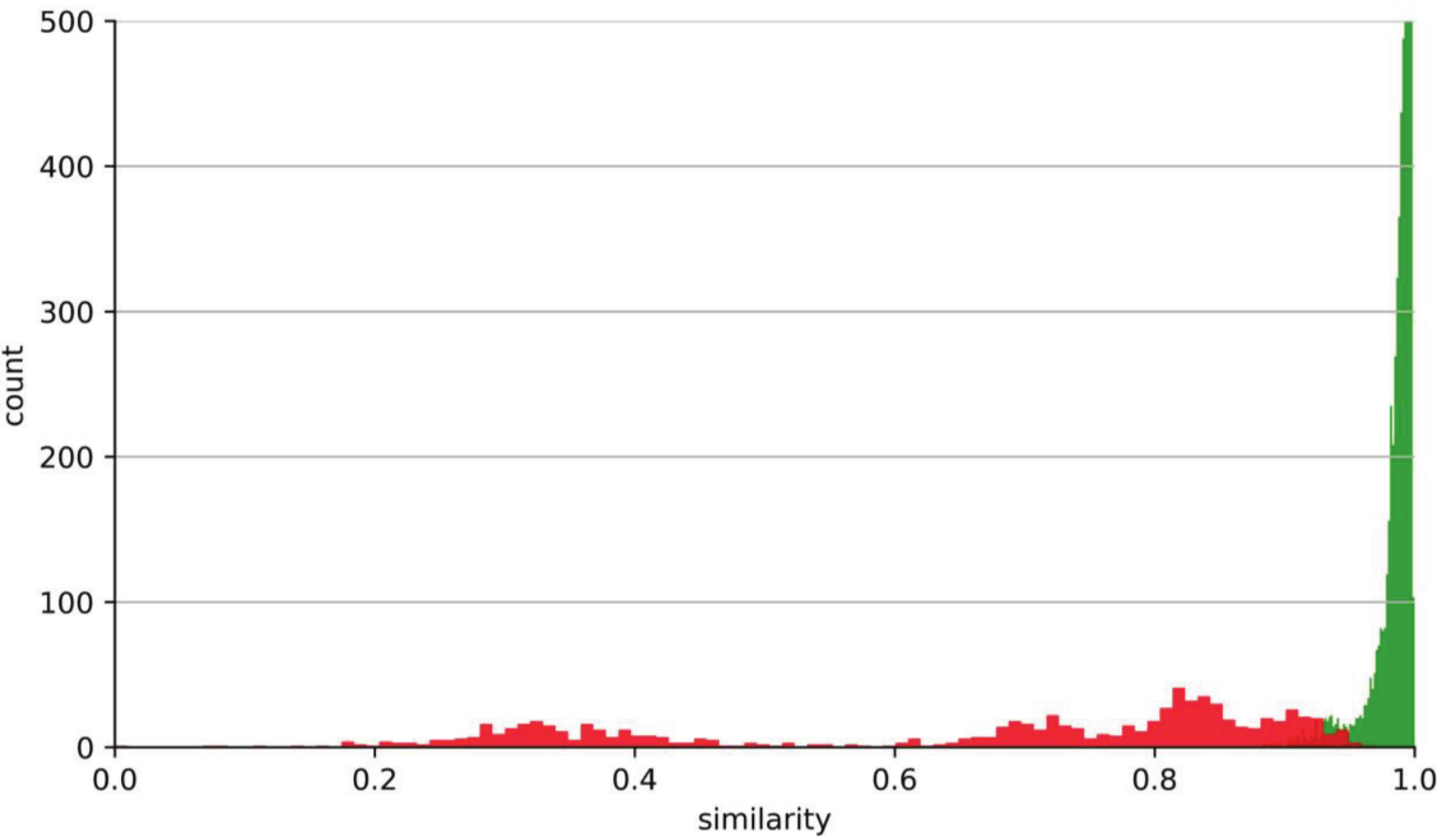}}
	\subfigure[``8'']
	{\includegraphics[width=0.3\linewidth]{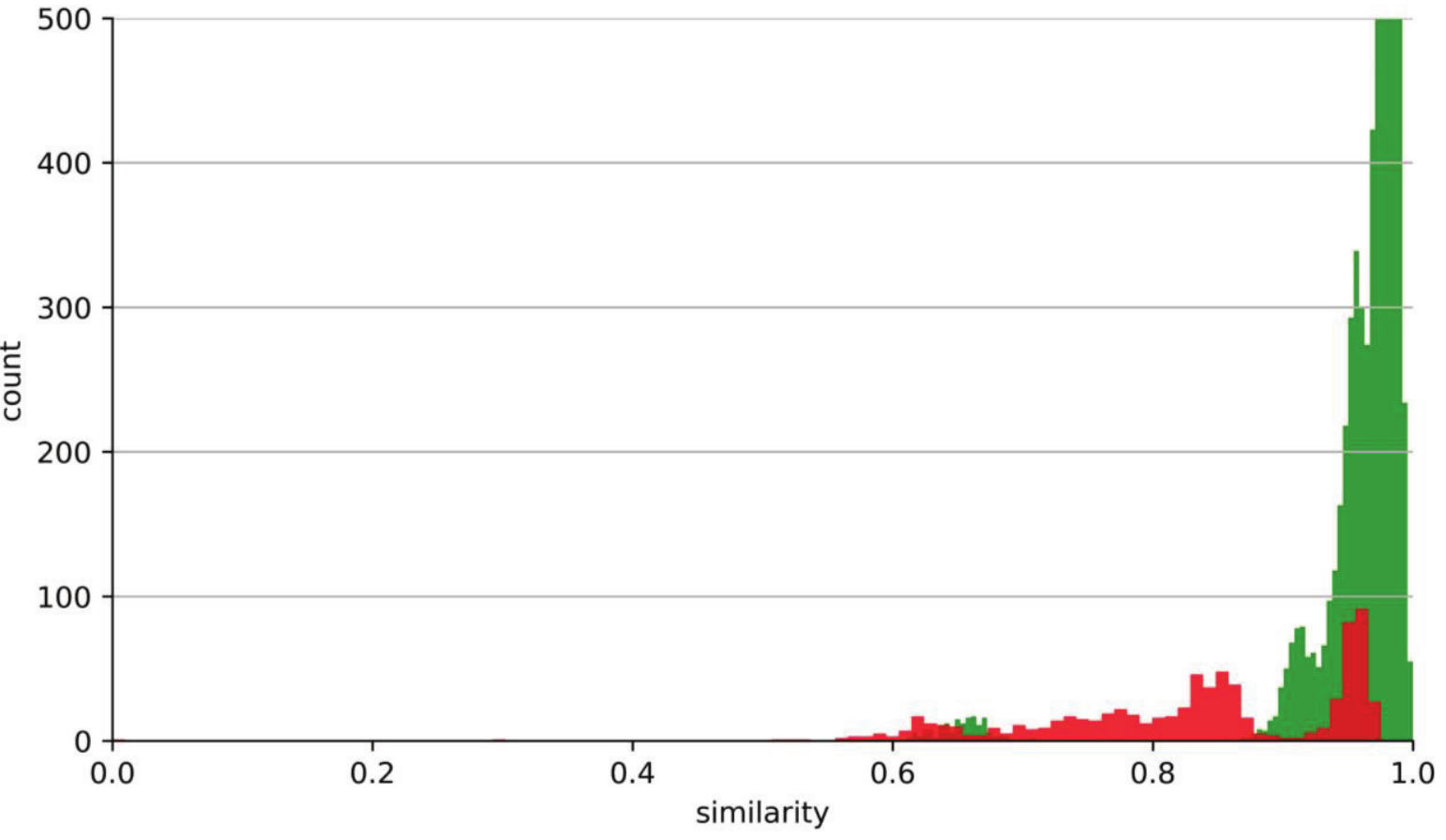}}
	\subfigure[``9'']
	{\includegraphics[width=0.3\linewidth]{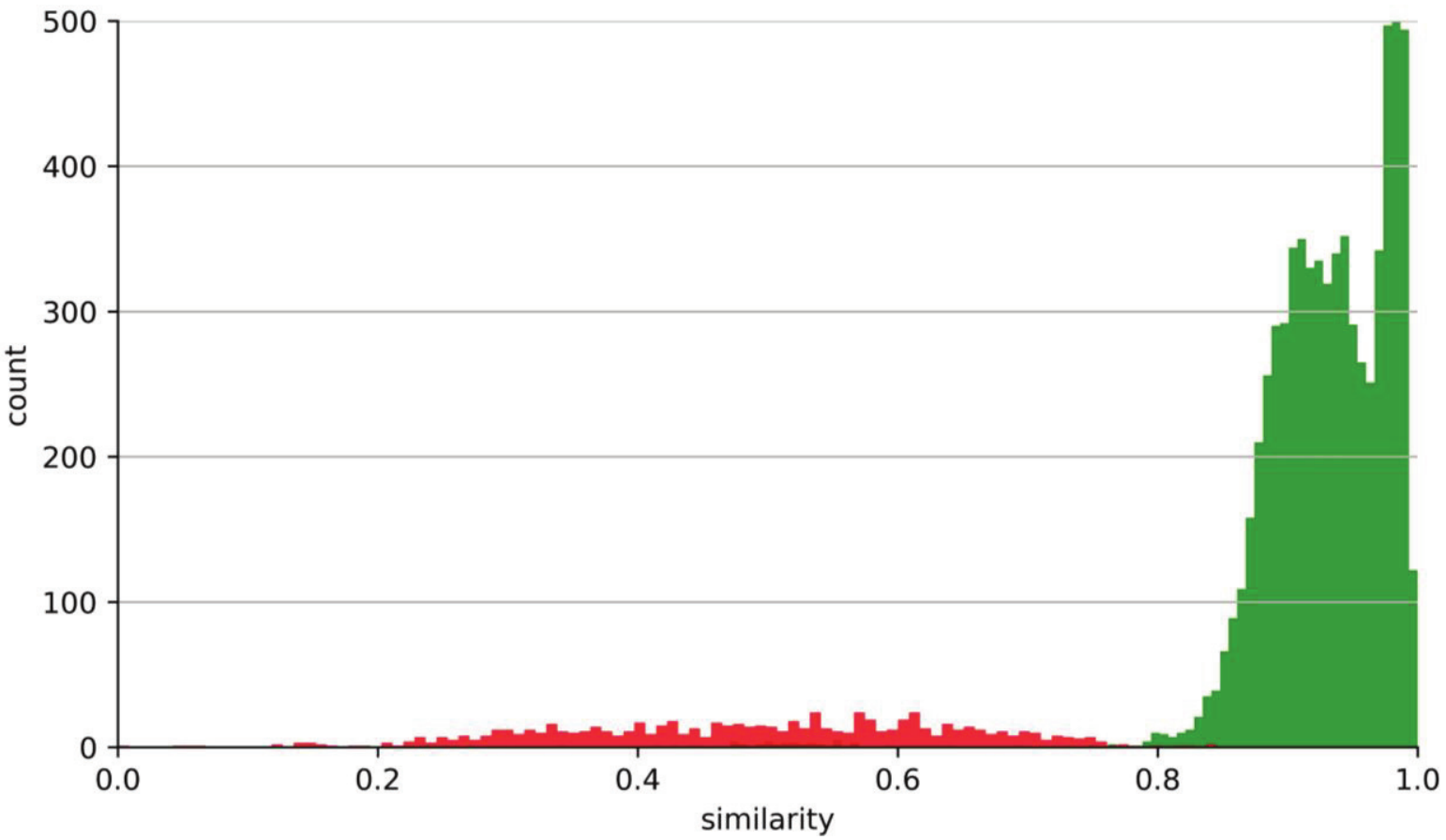}}
	
	\caption{\footnotesize{Feature difference distribution of ten classes in MNIST. The horizontal axis is the difference value and the vertical axis is the number of samples in each bin.}}
	\label{MNIST_hist}
\end{figure*}

\begin{figure*}[!htp]
	\centering
	\subfigure[T-shirt]
	{\includegraphics[width=0.3\linewidth]{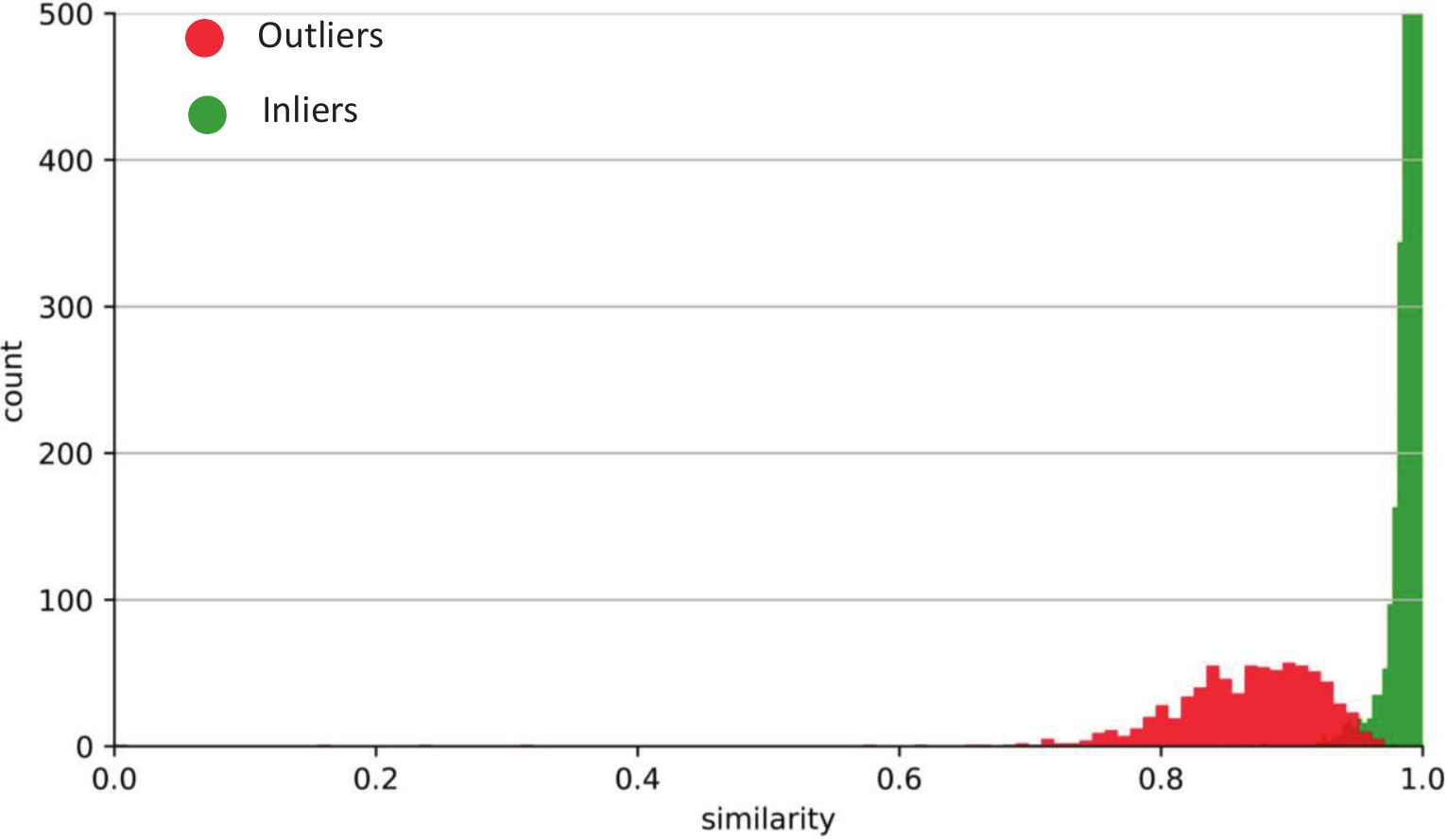}}
	\subfigure[Trouser]
	{\includegraphics[width=0.3\linewidth]{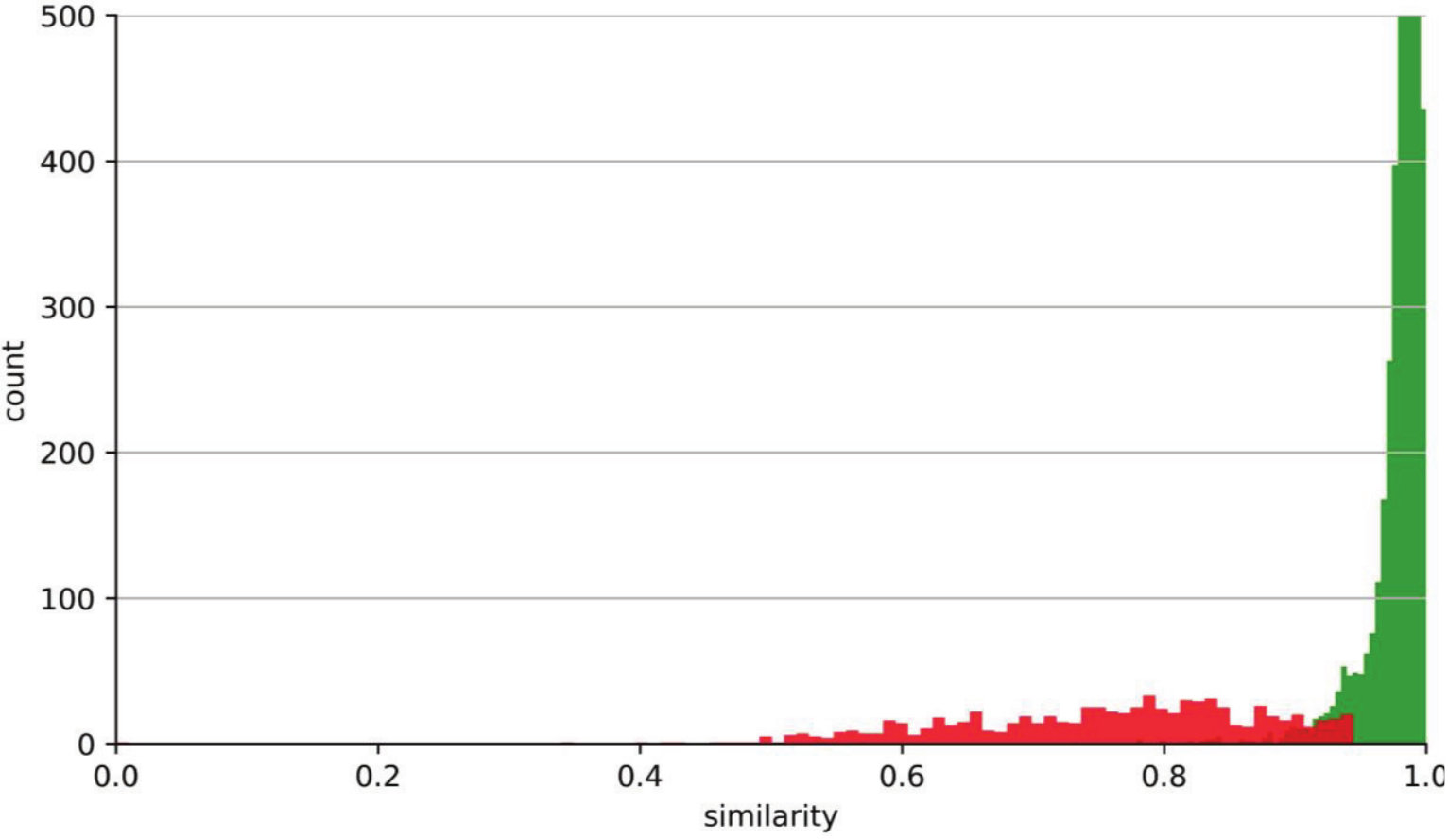}}
	\subfigure[Pullover]
	{\includegraphics[width=0.3\linewidth]{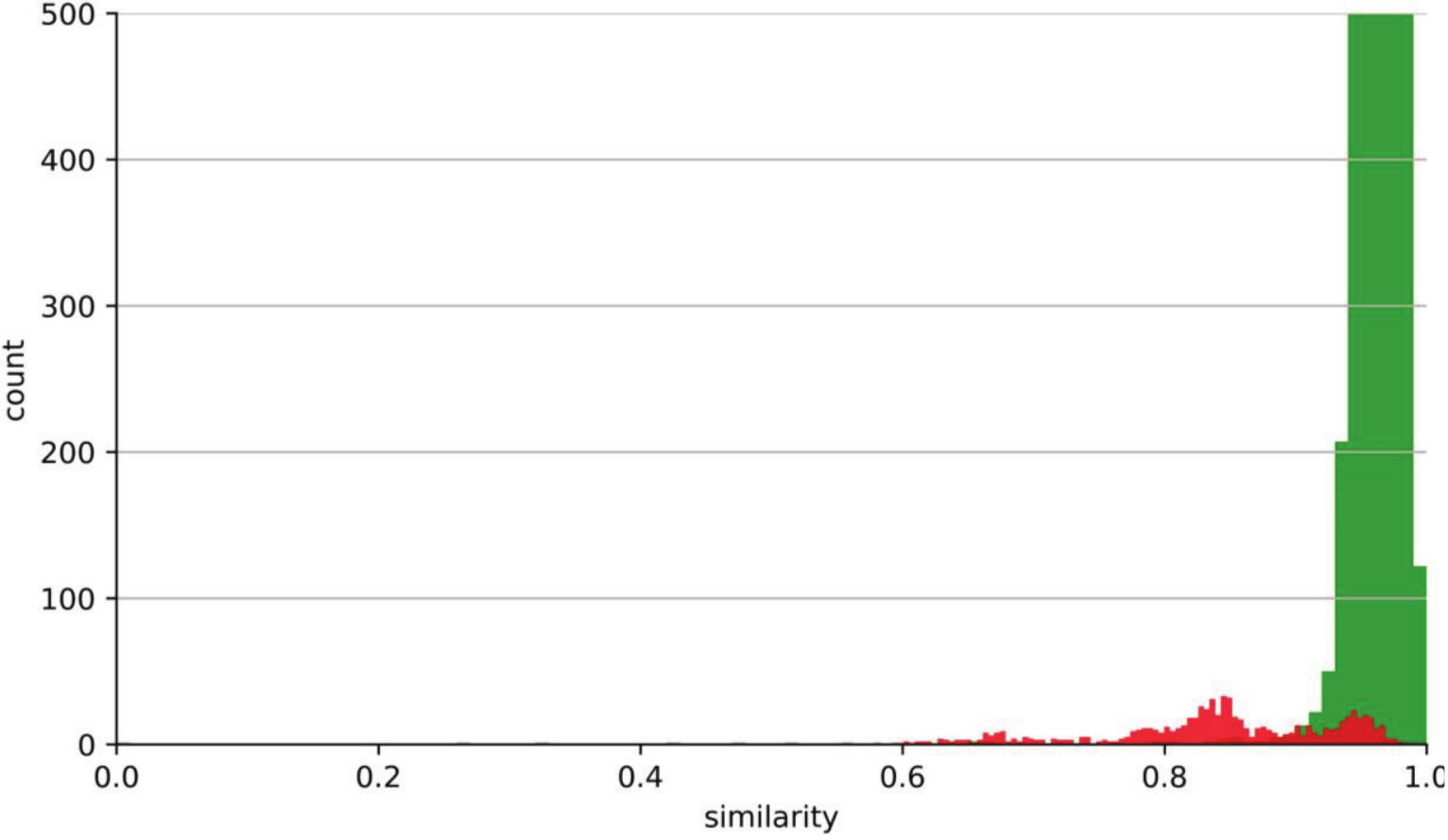}}
	\subfigure[Dress]
	{\includegraphics[width=0.3\linewidth]{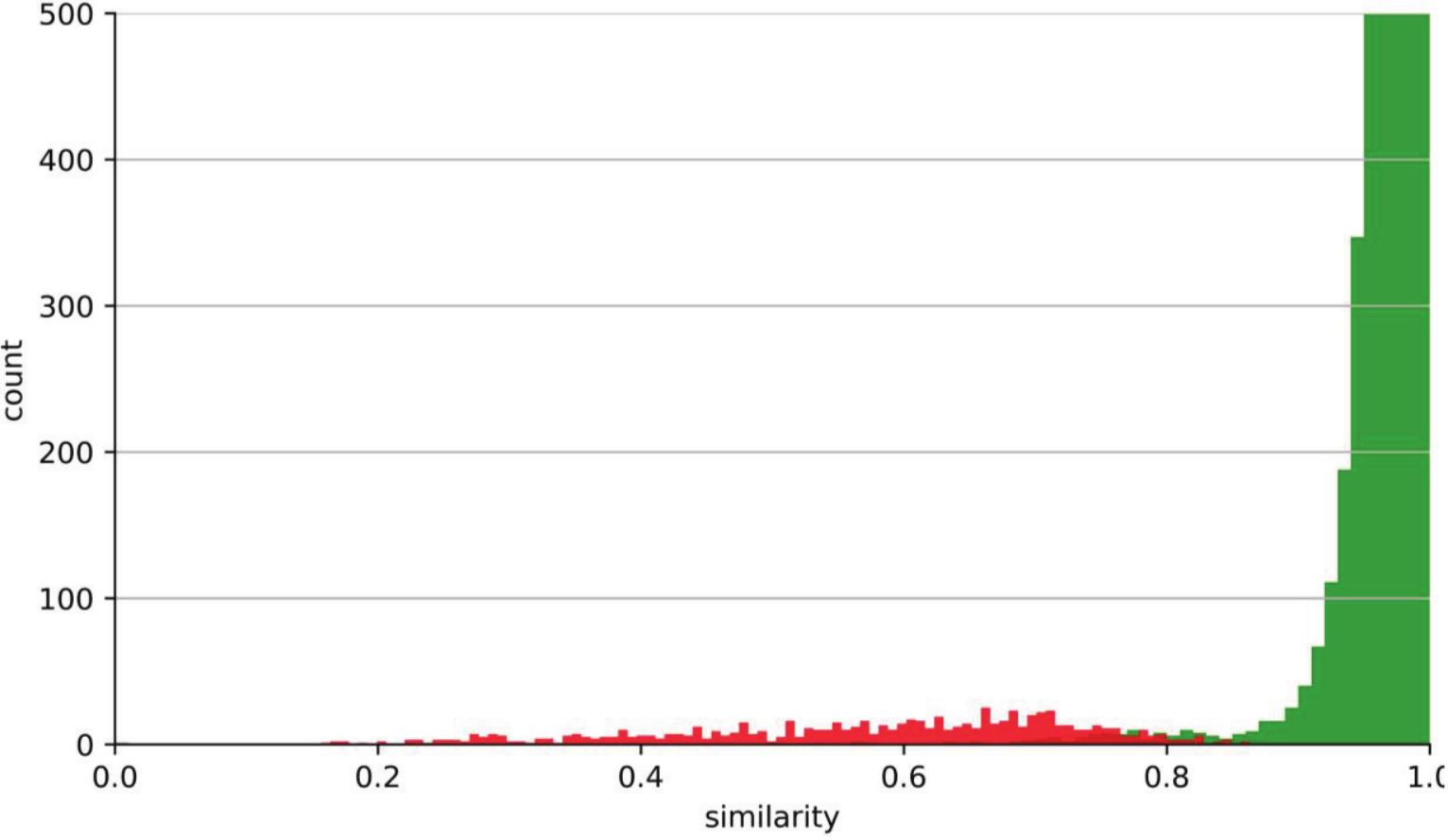}}
	\subfigure[Coat]
	{\includegraphics[width=0.3\linewidth]{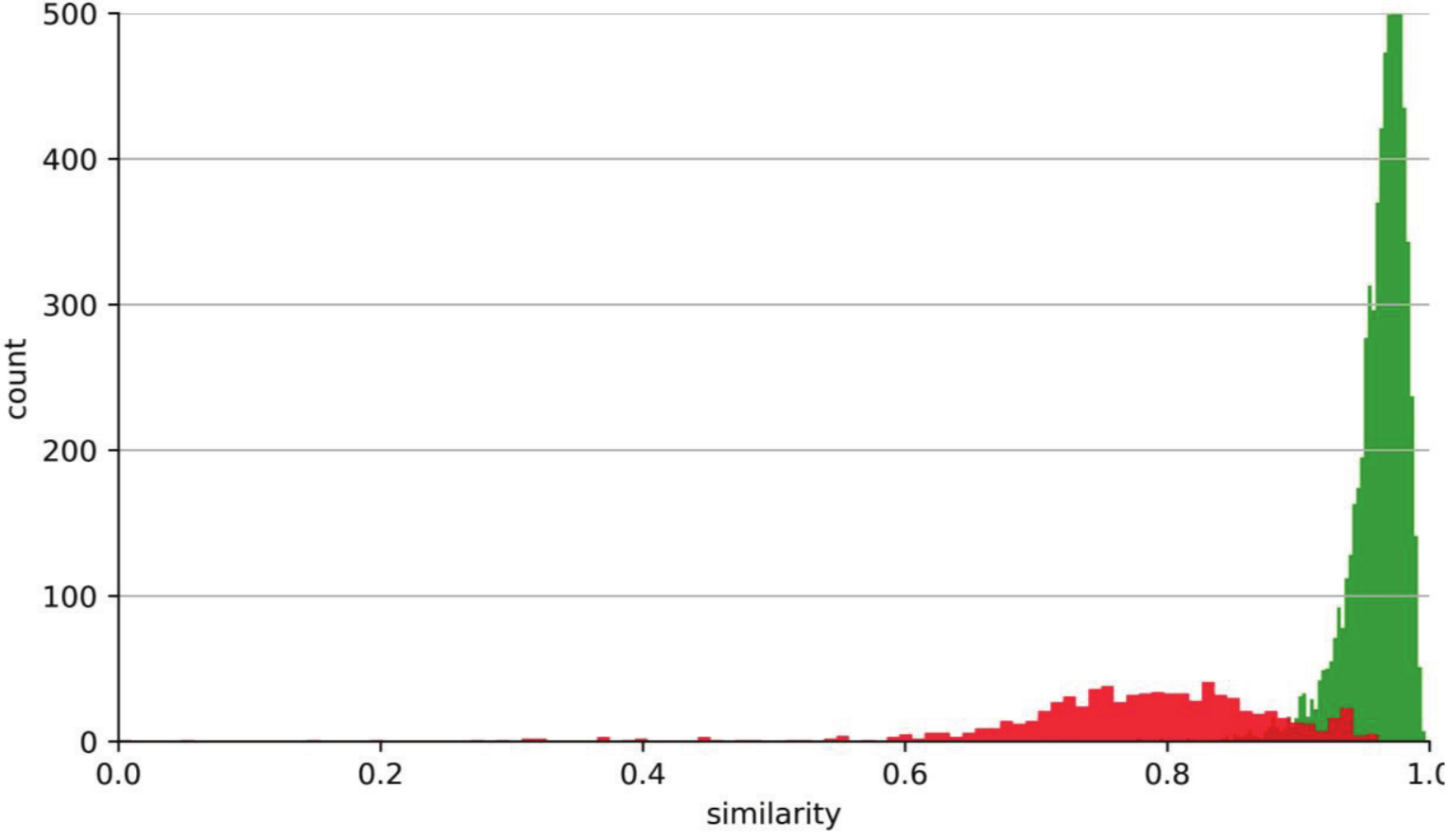}}
	\subfigure[Sandal]
	{\includegraphics[width=0.3\linewidth]{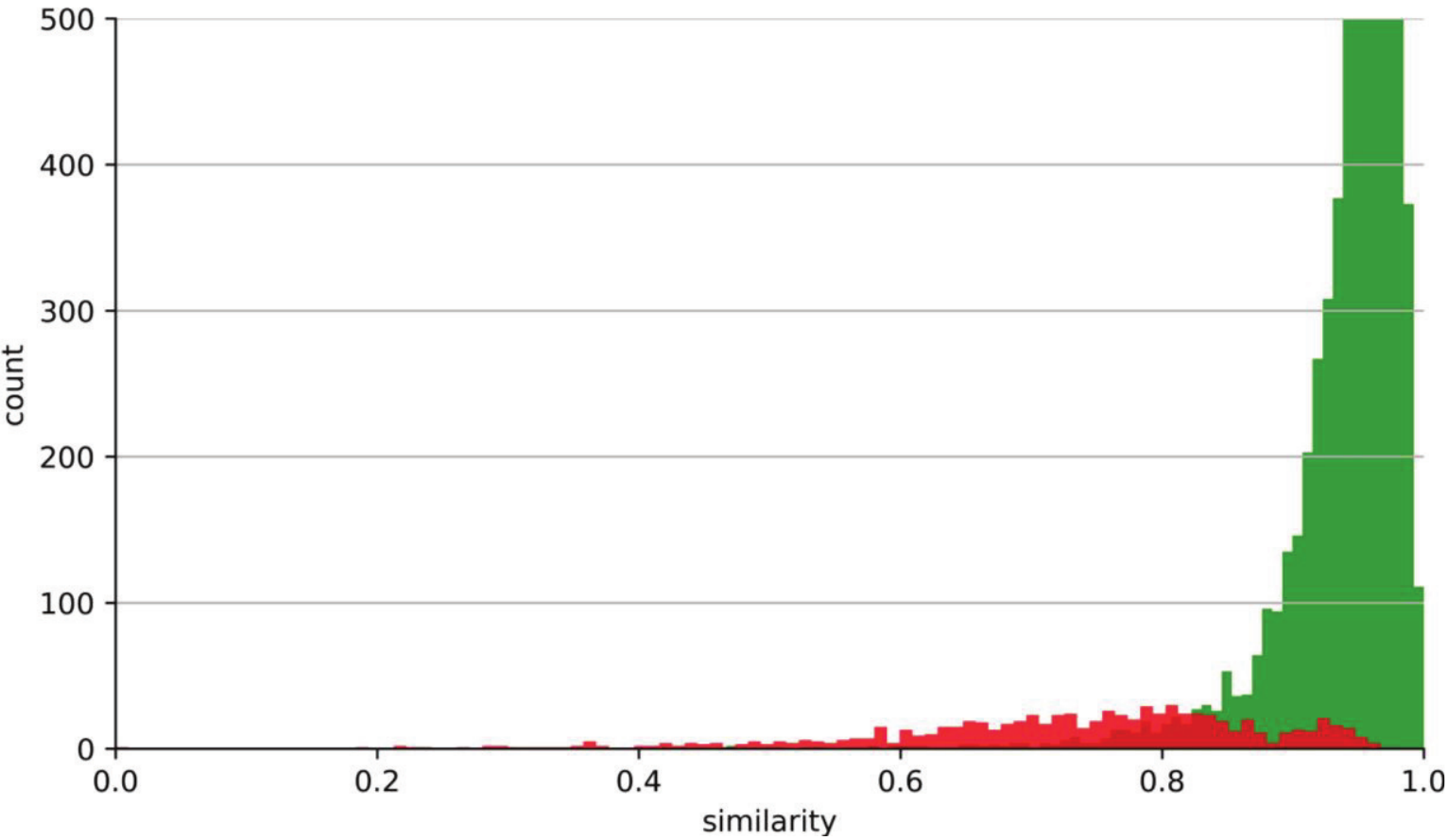}}
	\subfigure[Shirt]
	{\includegraphics[width=0.3\linewidth]{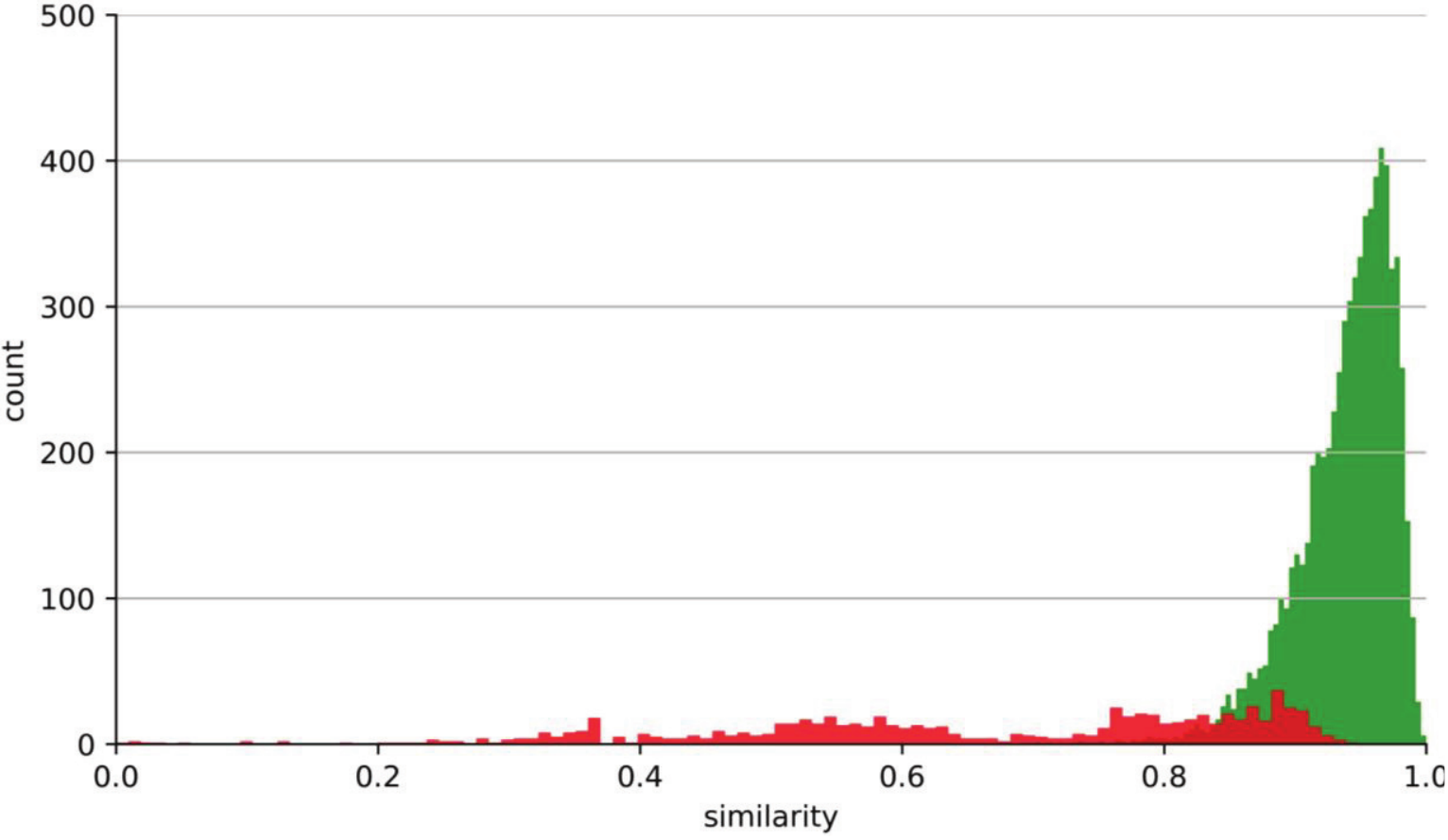}}
	\subfigure[Sneaker]
	{\includegraphics[width=0.3\linewidth]{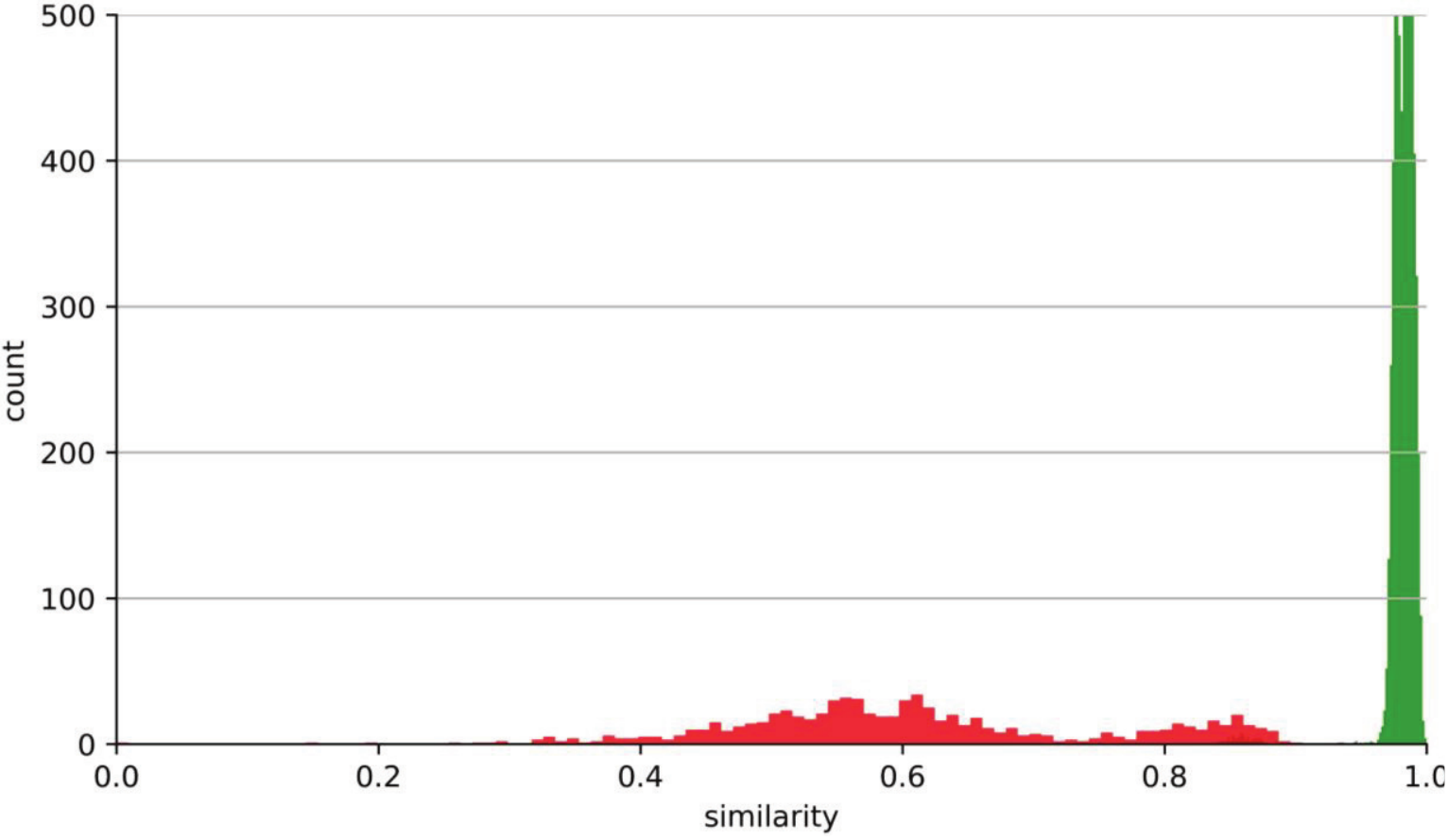}}
	\subfigure[Bag]
	{\includegraphics[width=0.3\linewidth]{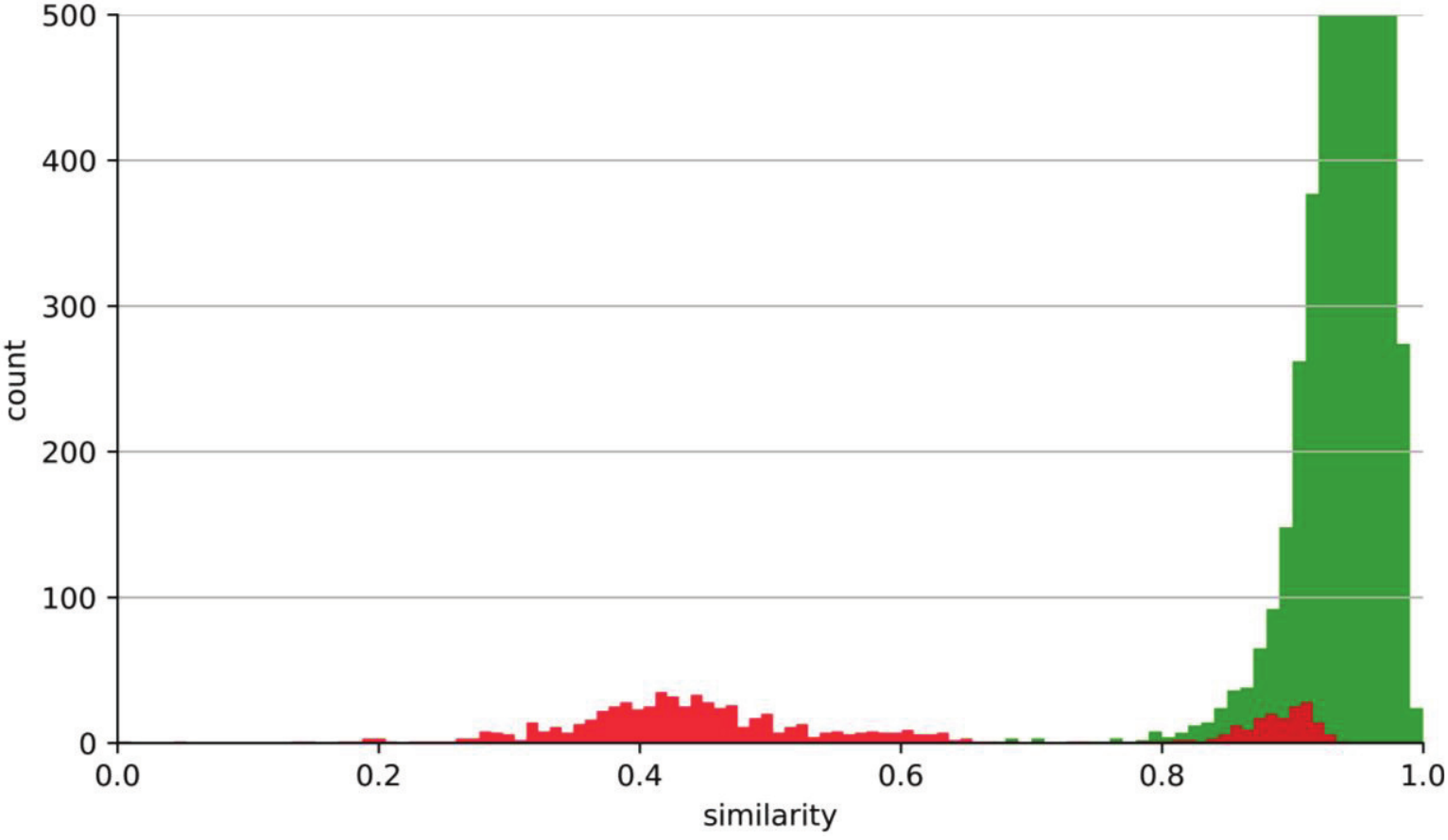}}
	\subfigure[Ankle-boot]
	{\includegraphics[width=0.3\linewidth]{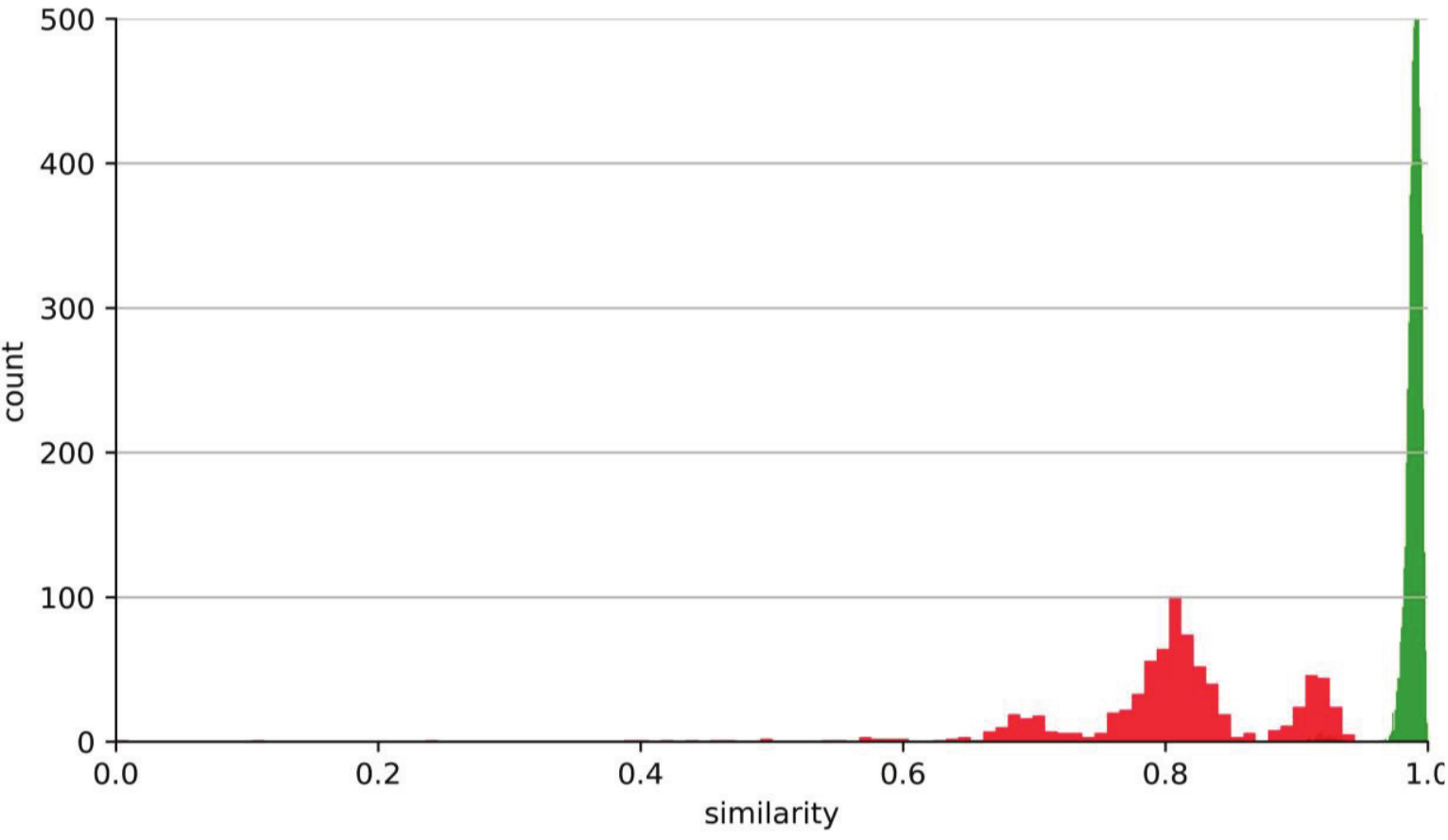}}
	
	\caption{\footnotesize{Feature difference distribution of ten classes in F-MNIST. The horizontal axis is the difference value and the vertical axis is the number of samples in each bin.}}
	\label{FMNIST_hist}
\end{figure*}

To illustrate that the learned features can be utilized to detect outliers, we calculate the differences between the extracted features of each inlier/outlier sample and the corresponding prototype read from memory defined in Eq. 9, and normalized the difference in a range $[0,1]$. Bigger value means a smaller difference. The histogram of the difference distribution is plotted in Fig. \ref{MNIST_hist} and Fig. \ref{FMNIST_hist}, which have $100$ bins. Each bin represents the number of outliers and inliers who have similar difference value. Inliers are shown in green bins and outliers are shown in red bins.

We can see an the obvious different distribution of outliers and inliers, which illustrates that it is feasible to detect outliers in the feature space. Moreover, one can see the distribution of outliers has a large variation, but the variation of inliers is rather small. The reason is that the outliers are from different classes, different from the memory prototypes. Thanks to the consistency learning, the learned features of inliers have a pretty high similarity.  

\section{TSNE visualization on CIFAR10 and CIFAR100}

The visualization of learned features on CIFAR10 and CIFAR100 are shown in Fig.~\ref{TSNE}. Specifically, (a) shows the visualizations of t-SNE of CIFAR10. $10$ classes are given according to top-to-bottom and left-to-right order. (b) shows the visualizations of t-SNE of CIFAR-100. $20$ classes are given according to top-to-bottom and left-to-right order. Blue dots are inliers and red dots are outliers. In most cases, outliers can be well separated from inliers.

\begin{figure*}[!htp]
	\centering
	\subfigure[CIFAR10]
	{\includegraphics[width=0.9\linewidth]{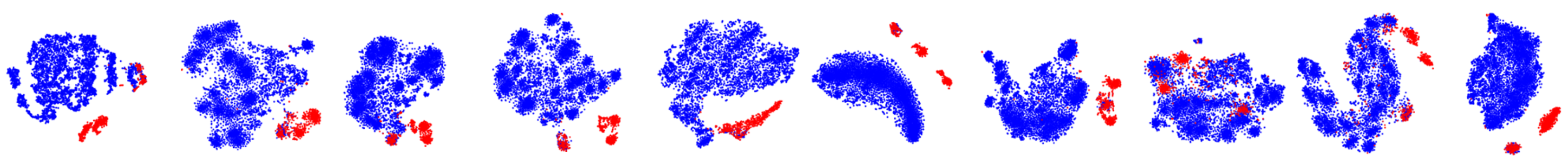}}
	\subfigure[CIFAR100]
	{\includegraphics[width=0.9\linewidth]{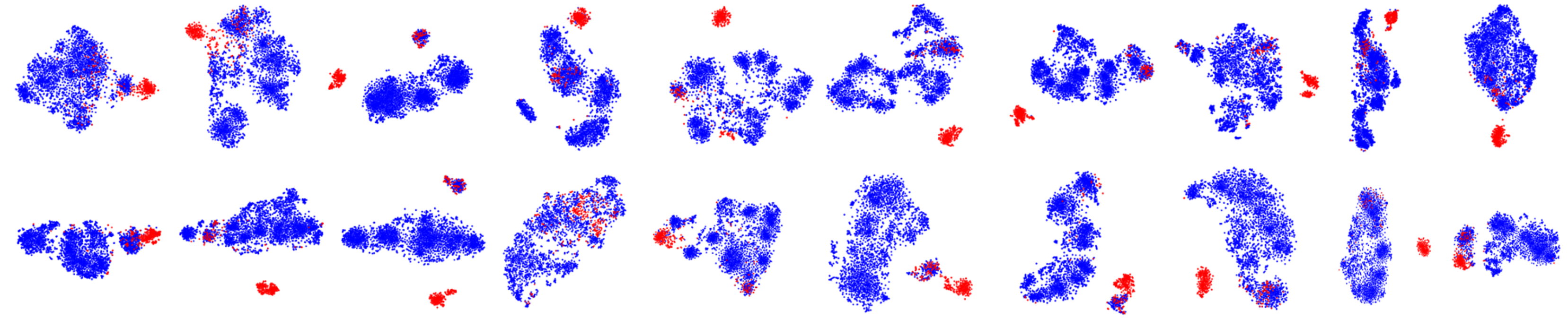}}
	\caption{\footnotesize{T-SNE visualization of features $f$ on CIFAR10 (a) (from class $1$ to class $10$) and CIFIAR100 (b) (from class $1$ to class $20$). Blue and red dots represent inliers and outliers in feature space. }}
	\label{TSNE}
\end{figure*}
\end{document}